\documentclass[pdflatex,sn-mathphys-num]{sn-jnl}

\usepackage{graphicx}%
\usepackage{multirow}%
\usepackage{amsmath,amssymb,amsfonts}%
\usepackage{amsthm}%
\usepackage{mathrsfs}%
\usepackage[title]{appendix}%
\usepackage{xcolor}%
\DeclareRobustCommand{\mjc}[1]{#1}
\usepackage{textcomp}%
\usepackage{manyfoot}%
\usepackage{booktabs}%
\usepackage{algorithm}%
\usepackage{algorithmicx}%
\usepackage{algpseudocode}%
\usepackage{listings}%

\theoremstyle{thmstyleone}%

\theoremstyle{thmstyletwo}%

\theoremstyle{thmstylethree}%

\begin{document}

\textbf{GOVERNMENT LICENSE}
 
The submitted manuscript has been created by UChicago Argonne, LLC, Operator of Argonne
National Laboratory (``Argonne''). Argonne, a U.S. Department of Energy Office of Science laboratory, is operated under Contract No. DE-AC02-06CH11357. The U.S. Government retains for
itself, and others acting on its behalf, a paid-up nonexclusive, irrevocable worldwide license in
said article to reproduce, prepare derivative works, distribute copies to the public, and perform
publicly and display publicly, by or on behalf of the Government. The Department of Energy will
provide public access to these results of federally sponsored research in accordance with the DOE
Public Access Plan. http://energy.gov/downloads/doe-public-access-plan.
\newpage

\title[ChatAtom: A Multi-Agent LLM System for MD]{ChatAtom: A Multi-Agent LLM System for Molecular Dynamics Simulation and Visualization}









\author*[1]{\fnm{Trupti} \sur{Mohanty}}\email{\mjc{tmohanty@anl.gov}}

\author*[1]{\fnm{Ming} \sur{Du}}\email{\mjc{mingdu@anl.gov}}
\author[2]{\fnm{Mitchell A.} \sur{Wood}}
\author[3]{\fnm{Brad} \sur{L. Boyce}}
\author[3]{\fnm{R\'{e}mi} \sur{Dingreville}}

\author*[1]{\fnm{Mathew J.} \sur{Cherukara}}\email{\mjc{mcherukara@anl.gov}}

\affil*[1]{\orgname{ Argonne National Laboratory}, \orgaddress{ \city{Lemont}, \postcode{60439}, \state{IL}, \country{USA}}}

\affil[2]{\orgname{Sandia National Laboratories}, \orgaddress{ \city{Albuquerque}, \postcode{87185}, \state{NM}, \country{USA}}}

\affil[3]{\orgname{Center for Integrated Nanotechnologies, Sandia National Laboratories}, \orgaddress{ \city{Albuquerque}, \postcode{87185}, \state{NM}, \country{USA}}}


\abstract{Molecular dynamics (MD) simulation resolves atomic-scale motion and 
predicts the structural, mechanical, and dynamic properties of materials, 
making it central to materials design and discovery. Yet an MD study requires an 
expert-driven workflow: designing the simulation, writing reliable input scripts, 
executing them, and analyzing the results with separate tools.  
Here we present ChatAtom, 
a multi-agent large language model system that translates a natural-language request 
into a complete MD workflow without the user writing input scripts or analysis code. 
Each agent has a defined responsibility within the workflow. An orchestrator plans 
the study; the LAMMPS and OVITO agents each use a generator--reviewer pair to prepare 
and review simulation and visualization scripts before execution; a high-performance 
computing agent dispatches jobs to remote resources; and an analysis agent interprets 
the completed run. The workflow is accessible through a browser interface that streams 
thermodynamic data and rendered frames during local execution, allowing users to 
follow the simulation as it progresses. The architecture also supports new capabilities 
through loadable skills without modifying the agents, enabling autonomous hypothesis generation 
and testing. We evaluate ChatAtom through script-generation benchmarks, an end-to-end study 
of crack propagation, and a closed-loop campaign. By bringing script generation, execution, 
visualization, and analysis into one reviewed and observable pipeline, ChatAtom lowers the 
barrier to conducting MD studies and broadens access to advanced simulation workflows.}


\enlargethispage{3\baselineskip}
\maketitle

\section{Introduction}
Molecular dynamics (MD) simulation is a powerful computational technique for tracking
atomic-scale motion and predicting the structural, mechanical, and dynamic properties of
materials, playing a central role in materials design and
discovery~\cite{tuckerman2000understanding}. It provides insight into processes that are
difficult or impossible to observe directly in experiment, such as how defects form and
move, how materials deform and fracture, and how they melt or change phase. Running an
MD study, however, requires a multi-step workflow: designing the simulation, choosing an
interatomic potential and a thermodynamic ensemble appropriate to the material, writing
an error-free input script in a domain-specific language such as that of
LAMMPS~\cite{lammps2022}, executing the job, and finally analyzing the resulting
trajectories in a separate tool such as OVITO~\cite{ovito}. These steps demand
substantial time and expertise, which limits the rapid exploration of scientific
hypotheses and presents a steep learning curve for non-expert users.

Large language models (LLMs) have advanced substantially in recent years, particularly
in scientific reasoning, code generation, information extraction, literature synthesis, 
data analysis, and interaction with specialized computational tools. A model connected 
to domain-specific tools can plan a task, invoke a tool, interpret its output, and decide 
what to do next, rather than responding in a single pass from memory. Acting through tools, 
and using the result of one action to guide the next, is what distinguishes an agent from a 
conventional chat interface. Systems built on this paradigm are now being reported across the 
sciences. ChemCrow routes a model through a set of expert-designed chemistry tools, so that 
the reasoning is performed by the model while the underlying computations are delegated to 
specialized tools~\cite{mbran2024chemcrow}, whereas Coscientist couples a model to web 
search, documentation, and a robotic platform to design and execute 
chemical experiments~\cite{boiko2023autonomous}. In materials simulation, 
DREAMS uses a planner and domain agents to coordinate density functional
theory workflows~\cite{wang2025dreams}.

Recent work has extended agentic workflows to atomistic simulation, with approaches
differing in the degree of control assigned to the language model. We classify the current approaches into two families. For the first family of methods, 
the LLM plays a central role in constructing and executing the simulation workflow. 
For example, a multi-agent framework has been developed in which specialized, task-specific 
agents coordinate structure preparation, interatomic potential selection, LAMMPS simulation, 
and per-property analysis~\cite{vriza2026multiagentic}. A related system uses an LLM agent to select interatomic 
potentials and generate and execute simulation scripts~\cite{somasundaram2026ursa}, 
while other efforts extend agentic workflows to broader materials simulation~\cite{liu2026masgent} 
and biomolecular workflows~\cite{ma2026mdagent}. A separate line of work targets 
generation quality itself, through domain-specific post-training~\cite{shi2026mdagent2} 
or iterative optimization using reward signals~\cite{wang2026mdforge}. 
A second family restrict the LLM primarily to planning and decision 
making while deterministic software performs simulation and validation. 
El Agente Potente, for example, uses the model for planning and routing within a typed, 
provenance-aware execution graph~\cite{ko2026elagente}, whereas Agent-MD uses the model only to plan a campaign (a sequence of related runs) and
to review events the rules cannot resolve, leaving the runs themselves to deterministic
code~\cite{wang2026agentmd}. Alongside these developments, 
evaluation studies have begun to examine the correctness of LLM-generated LAMMPS 
workflows~\cite{holbrook2026evaluating}.

These systems show that a natural-language request can drive an atomistic
simulation workflow, but several aspects can be taken further. For example,
 in \cite{vriza2026multiagentic}, the simulation and analysis are 
 orchestrated with a static workflow involving a series of agents. 
 This design offers predictability, but sacrifices the flexibility 
 of the agent system, limiting its capability of responding to arbitrary 
 user requests. Additionally, visualization is produced by fixed, provisioned routines, so a
user who wants a particular view, such as special atom color codes or isolated defect display, 
needs to create the visualization routines themselves. User interactions 
\cite{vriza2026multiagentic,wang2026mdforge} are also primarily terminal-based,
and simulations are not streamed as they run, so users can inspect 
the results only after execution is complete.

Here, we present ChatAtom, a multi-agent LLM system that turns a
natural-language request into a complete MD workflow without the user writing simulation
code. Its agents are defined by role: an orchestrator responds to general user requests and plans the
study, a LAMMPS agent and an OVITO agent write the simulation script and the
visualization pipeline, a high-performance computing (HPC) agent dispatches jobs to
remote resources, and an analysis agent interprets the completed run. 
Each script-writing agent pairs a generator with a reviewer, so that
errors detected by the reviewer are corrected before execution, and the
visualization pipeline is written for the request rather than chosen from a fixed set.
ChatAtom features a design combining the determinism of fixed, static graphs, and the flexibility of tool-calling agents: the user-facing orchestrator can adaptively handle user requests as a tool-calling agent, while the LAMMPS agent and OVITO agent are concatenated in a deterministic workflow. 
ChatAtom is accessed through a browser interface, where locally executed 
simulations can be observed through streaming thermodynamic data and rendered frames.
Human-in-the-loop control keeps the plan and the job submission with the user, 
and every run is saved to a local run directory for subsequent inspection and analysis.
We evaluate ChatAtom on a benchmark of eight prompts issued to six frontier models, where
it raises the number of executable scripts from 92 of 144 under zero-shot prompting to
132. For the three tasks with published reference values, the number returning a correct
result rises from 25 to 50 of 54.

We further extend ChatAtom beyond single simulations to hypothesis generation and
evaluation. Materials behavior emerges from competition among mechanisms, and exploring
the conditions that favor one outcome over another requires automated simulation
campaigns~\cite{boyce2026materials}. Coupling AI-driven hypothesis generation with agentic
systems opens a realistic path toward end-to-end automated
discovery~\cite{jacobsson2026hypotheses}, and recent work in materials research has
begun to generate and screen candidate
hypotheses~\cite{kumbhar2025hypothesis,zhou2026autonomous,roy2026autonomous}. In ChatAtom
the capability is supplied as a loadable skill rather than as a change to the agents.
The user states an objective; the orchestrator proposes the simulation workflow and the
range over which each parameter may vary; the script is generated once and held fixed as
a template, so that successive runs differ only in the parameters substituted into it;
and the objective metric is computed from the simulation output, so that each trial is
proposed from the parameters and results of the ones before it. We demonstrate this workflow on
simulations of crack growth in copper over a parameter space of temperature, strain rate, and
crack length, where the system proposes its own runs and carries them out unattended.
Taken together, these capabilities lower the barrier to MD studies and broaden access to
advanced simulation workflows for a wider community of researchers.

\section{Methods}
\label{sec:methods}

\begin{figure}[h]
\centering
\includegraphics[width=0.9\textwidth]{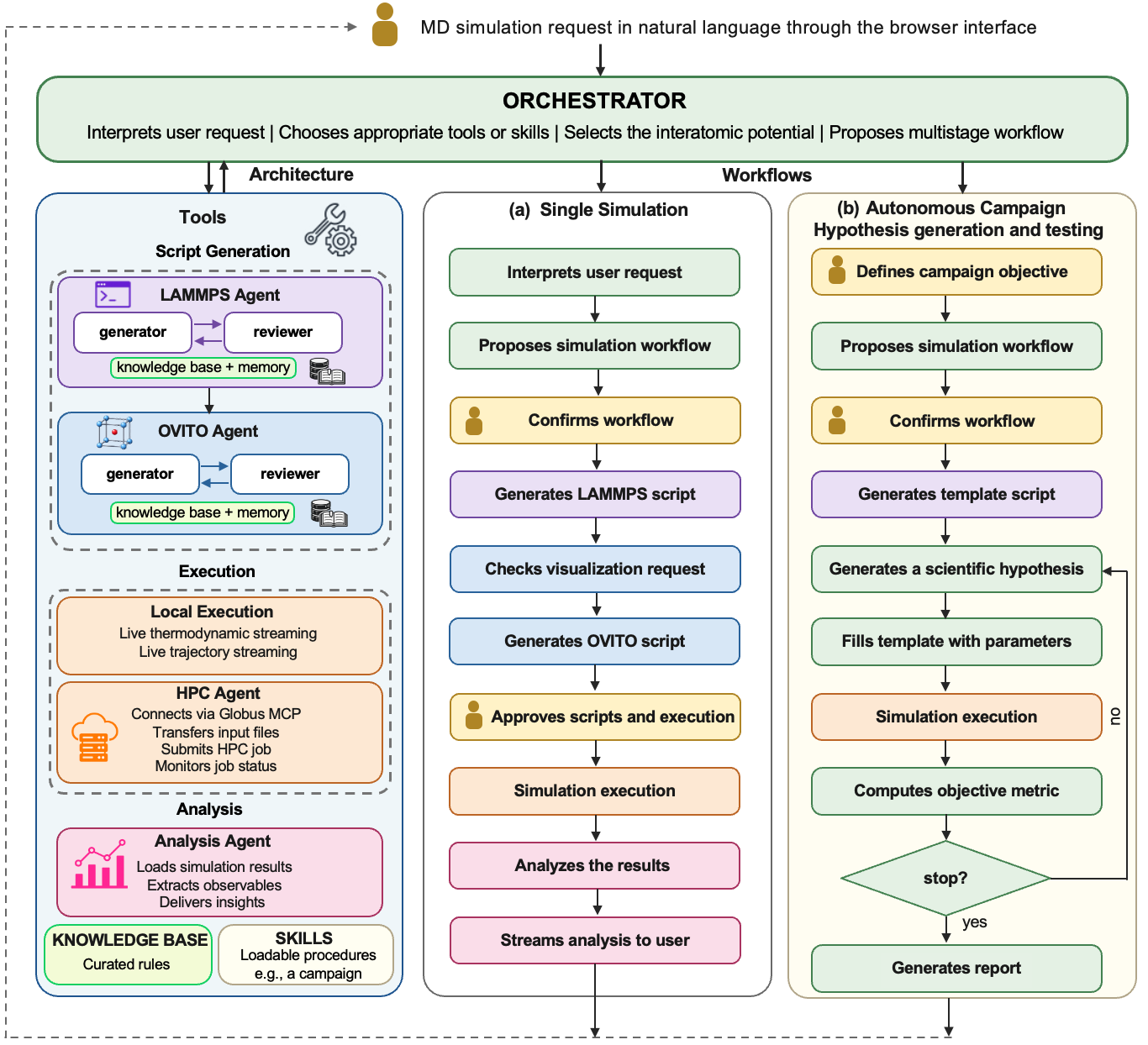}
\caption{ChatAtom architecture and workflows. The orchestrator serves as the main
user-facing agent, interpreting user queries and invoking appropriate tools and skills. 
Script generation is delegated to a specialized tool consisting of a LAMMPS agent and an OVITO
agent, each of which pairs a generator with a reviewer and draws on a shared knowledge
base and persistent memory; Simulations are executed either locally, with real-time streaming of thermodynamic 
data and trajectory frames, or on HPC systems through the HPC agent. The analysis 
agent extracts observables and delivers insights.
The workflows illustrate (a) a single-simulation workflow, in which the orchestrator 
selects an interatomic potential, develops a multistage simulation plan, coordinates 
scripts generation, executes the simulation, invokes analysis agent, and (b) an autonomous 
campaign, in which the agent iteratively generates hypotheses, executes simulations, 
and evaluates objective metrics in a closed loop. In both workflows, the color of each 
step matches that of the corresponding component in the architecture. 
The distinct background color of the autonomous campaign indicates that it is provided as a loadable skill.
.}\label{fig1}
\end{figure}

ChatAtom is implemented as a multi-agent system in which a top-level orchestrator coordinates 
specialized agents responsible for different stages of the MD workflow, including script 
generation, execution, visualization, and analysis. The flexible workflow also supports 
autonomous hypothesis generation and testing as a loadable skill that can be invoked 
when needed. Figure~\ref{fig1} summarizes the overall architecture, 
and the following subsections describe each component in detail. The orchestration layer and LLM communication handling are implemented using LangGraph \cite{langgraph} and DSPy \cite{dspy}.

\subsection{Orchestrator}

The orchestrator serves as the user-facing conversational agent and the single point of entry to ChatAtom. 
All user requests are routed through the orchestrator, including requests to initiate a 
simulation, analyze a completed run, or ask general questions 
related to molecular dynamics. It is implemented as a tool-calling 
agent using LangGraph~\cite{langgraph}. At each interaction, 
the model selects from the available tools based on their descriptions, 
executes the selected tool when required, interprets the returned result, 
and responds directly to the user.

For a new simulation request, the orchestrator first constructs a simulation plan rather
than proceeding directly to script generation. It replies with a complete plan, stating
the material, the system size, the interatomic potential, the boundary conditions, the temperature, the ensemble,
the duration of each phase, the frame sampling, and any visualization preference the
request implies.

Every simulation uses an interatomic potential from OpenKIM, the Open Knowledgebase of
Interatomic Models~\cite{openkim}. The orchestrator looks into a local registry, a text
file of the OpenKIM models available for different elements, which can be extended with
new entries. A model is therefore proposed from the registry rather than from the
language model's own memory, and a model specified by the user in the conversation takes
precedence over the proposal. Before the simulation starts, the chosen model is
installed through the KIM API if it is not already present.

The orchestrator's proposed plan is open to revision: the user may ask for changes over
several turns, and the orchestrator updates the plan and asks again, as shown in Figure~S1 of the supplementary
information. Once the user confirms the plan, the orchestrator calls a tool that encapsulates a deterministic workflow composed of the LAMMPS agent, which generates the input script, and the OVITO agent, which writes the visualization script, to create the simulation script. Both scripts are then displayed in the interface, where the user may edit them,
and execution begins only on approval, either on the local machine or, through the HPC
agent, on a cluster. A request to interpret a finished run is passed to the analysis
agent. The flexible architecture also allows the orchestrator to be extended with new capabilities,
as demonstrated by autonomous hypothesis generation and testing in Section~\ref{sec:campaign-results}. This is implemented as a campaign in which each simulation is selected based on the results 
of previous runs. Rather than building a dedicated workflow, we provide the campaign procedure as a loadable skill, together with the tools it
requires.
The orchestrator loads this 
skill into its prompt only when a campaign is requested, avoiding the need to 
include campaign-specific instructions in every conversation. The campaign therefore 
reuses the same agents and script-generation process as an individual simulation, 
as described in Section~\ref{sec:campaign}. Throughout the workflow, the orchestrator maintains 
the conversation state, which is checkpointed to a local SQLite database so that a session can be resumed later.

The script generation process is packaged as a tool that the orchestrator can call. The tool is a LangGraph workflow consisting of the LAMMPS agent and the OVITO agent. 

\subsection{LAMMPS agent}
\label{sec:lammps}

The orchestrator passes the approved plan, with the full detail of the simulation, to
the LAMMPS agent, which writes the input script through a generator and a reviewer
working in an iterative loop. The generator first drafts the script from the plan, the knowledge
base, and the failures recorded in memory for earlier runs of the same material. The
reviewer then reads that draft against the same knowledge base rules and returns a
verdict, the errors it found, if any, and how to fix them. When a draft is rejected,
those errors and fixes are passed back to the generator, which writes a new script
from the plan, incorporating the suggested corrections, and a script is accepted only when the reviewer
reports no errors, within at most three attempts; an exchange of this kind is shown in
Figure~S2 of the supplementary information. Each attempt is also recorded in a
persistent memory as a success or a failure, so that the failures can serve as lessons
in later simulations.

Both generator and reviewer draw on a curated knowledge base, a fixed set of rules
supplied with every call rather than left to what the model has memorized. Each rule
states a condition the script must satisfy. For example, \texttt{kim init} must precede
\texttt{create\_box} and \texttt{kim interactions} must follow it, and a script that
departs from this order fails at startup; atomic masses must be set for each element,
because a KIM model does not supply them; a thermostatted run must be preceded by
\texttt{velocity all create} at the thermostat temperature, so that it does not begin
from a velocity distribution inconsistent with the ensemble. The generator follows these 
rules when writing the script, and the reviewer treats violations as grounds for rejection.

\subsection{OVITO agent}

Visualization is handled by a second generator and reviewer pair, the OVITO agent. 
It writes an ordered list of operations that OVITO applies to a loaded
trajectory: an operation might compute a per-atom quantity, select or remove atoms, or
color those that remain. OVITO runs them one after another on a frame and draws the
image. The agent is invoked only when a particular visualization is asked for in the
proposed plan; otherwise atoms are colored by particle type across all frames.

When a visualization is requested, the generator receives the request, the material, and
a list of the quantities the LAMMPS script has already computed. If that script ran
common-neighbor analysis per atom~\cite{stukowski2012structure}, for instance, the
values are already in the dump file and the pipeline reads them rather than computing
the analysis again. The knowledge base lists which modifiers exist and which module each
comes from, and sets the rule that governs frames: OVITO rebuilds each frame by running
the pipeline independently for that frame, so a property added outside the pipeline is lost at the next
evaluation. The reviewer rejects an unlisted import or a property created outside the
pipeline and returns the fixes with it, and a rejected script is rewritten over at most
three attempts, with earlier visualization failures read from memory beforehand and new
ones recorded after.

The accepted script is Python that ChatAtom executes itself, so it passes a validation
check in code before it runs: anything that touches the filesystem, starts a process or
evaluates code at run time is refused, and imports are limited to visualization and
numerical libraries. If a script fails that check, rendering
falls back to coloring by particle type. 

\subsection{HPC agent}

The HPC agent runs a simulation on a remote cluster rather than the local machine. Like
the orchestrator it is a tool-calling agent; it connects to a Globus Model Context
Protocol (MCP) server~\cite{globusmcp} and loads the tools it exposes for file transfer,
job submission and status queries. Given an approved simulation, the agent fills a batch
submission script from a fixed template for the cluster's scheduler, creates a remote
run directory, transfers the approved LAMMPS input and the batch script, and submits the
job. The scheduler returns a job identifier, which the agent keeps for the rest of the
session, so the user can ask for the status of that job at any point; once it completes,
the output files are transferred back and the run can be analyzed like any other.

\subsection{Analysis agent}

The orchestrator can invoke the analysis agent on a specified past simulation upon user
request. The analysis agent is a tool-calling agent with tools that include the
following capabilities: listing the recent runs, reading a run's log and loading its
trajectory into the viewer, computing a stress--strain curve from the thermodynamic
output, loading the trajectory frames and measuring dislocation density and Burgers
vector families in each of them through dislocation extraction
analysis~\cite{stukowski2012dxa}. The agent extracts the physical properties from the
simulation results using these tools and compiles them into a human-readable report,
which is returned in the interface alongside the thermodynamic plots and the rendered
trajectory.

\subsection{Human-in-the-loop control and execution}
\label{sec:execution}

ChatAtom checks with the user before it proceeds: once before a script is written, and
again before it is run. A simulation is not generated until the orchestrator has
proposed a complete plan and the user has accepted it, and the user may instead ask for
changes or drop the request. The script is then shown in the interface, where it can be
edited, and the edited version is the one that executes. The user also chooses whether
it runs on the local machine or on a cluster; on the HPC path the batch script is shown
in the same way, and nothing is transferred or submitted until the user approves it.

Once a script is approved, the run proceeds on the local machine, with its output
streamed to the interface as described in Section~\ref{sec:streaming}, or on a cluster
through the HPC agent. In either case it is given its own timestamped directory holding
the input script, the thermodynamic log, the trajectory dumps and the visualization
script, so that it can be revisited and analyzed later. A campaign is the exception to
this sequence, approved once rather than at every step: the user accepts the design, the
parameters to be varied with the bounds they may take, and the template script that
every trial will use, after which the loop runs on its own.

\subsection{Hypothesis generation and testing}
\label{sec:campaign}

Beyond single simulations, ChatAtom can generate and test its own hypotheses, running the
whole workflow inside a closed loop. Such a study, a campaign, is defined by a physical
objective to optimize, a template of fixed simulation settings shared by every trial,
and a set of tunable parameters with allowed ranges. Once the user
confirms that design, the orchestrator drives the loop through a dedicated set of
campaign tools.
While the present hypothesis testing is open-ended but constrained by a single objective definition, the collection of agents and their inter-dependencies of ChatAtom offers extensions to much broader application.

Each iteration follows a fixed sequence: the orchestrator reads the campaign state (which includes the plan it was given and the record of the trials completed so far), proposes a
hypothesis with a new set of parameter values, runs a simulation with those values under
the template, measures the objective from the resulting trajectory, records the trial,
and checks the stopping criteria. Each trial is stored together with the hypothesis that
produced it, so the reasoning behind a campaign can be read back afterwards.

The script is generated once and approved as a template, and later trials are produced
by substituting the new parameter values into it without a further pass through script
generation. The template fixes every setting that is not under test, among them the
system size, the potential, the sequence of phases, their durations and the random
velocity seed, so that a difference in outcome can be attributed to the parameters under
test. 

The loop ends when a stopping criterion is met: a set number of iterations, a target
value of the objective being reached, or three failed trials in succession. The user can
also pause or cancel a campaign while it is running. ChatAtom then generates a report that
plots the objective against iteration and against each parameter, tabulates every trial,
and writes a narrative interpretation into the report in place.

A tool written for the campaign computes the objective from the trajectory. In the
campaign presented in Section~\ref{sec:campaign-results}, it is the increase in the
crack surface area during loading, which is to be minimized during the campaign. The
parameter space is formed by the temperature, strain rate and the length of the initial
crack.

\subsection{Interface and live streaming of execution}
\label{sec:streaming}

The user interacts with ChatAtom in a browser, served by a Python backend; Section~S1 of
the supplementary information shows the interface. Each conversation is a persistent
session, checkpointed so that it can be resumed with its history and its simulation
state intact.

Both the agents' replies and the output of a running simulation reach the browser as
they are produced. For a simulation on the local machine, thermodynamic quantities are
sent point by point as LAMMPS logs them and plotted as they arrive, and each dump file
is passed to a standalone render process that applies the OVITO pipeline and returns the
rendered configuration tagged with its frame number. The frames are shown
in sequence, so the trajectory animates while it is being produced. LAMMPS runs as a separate process while a monitoring thread forwards its output and a
render thread handles the frames, so the interface stays responsive throughout. 

\section{Results and Discussion}
We evaluate ChatAtom in three ways. First, we compare the LAMMPS scripts generated by ChatAtom with those 
produced by the same underlying model when prompted directly. Second, we evaluate the complete workflow, 
from a single natural-language request to the final analysis. Third, we evaluate autonomous campaign 
operation by allowing the system to generate hypotheses, propose simulations, execute them, 
and assess the resulting outcomes.

\subsection{Script generation compared with zero-shot prompting}

\label{sec:benchmark}

ChatAtom is a multi-agent system in which different language models can be assigned to 
individual agents through the system configuration. For the comparisons reported here, 
however, all agents in a given configuration used the same frontier model, allowing 
ChatAtom to be compared directly with that model prompted independently. Script generation provides a common basis for 
this comparison, since both the standalone model and ChatAtom produce a LAMMPS input script 
from the same natural-language request. We evaluated three conditions: the model prompted 
zero-shot, the same model operating within ChatAtom, and ChatAtom with a set of LAMMPS scripting 
rules added to its knowledge base. The model was held fixed across all three conditions,
and persistent memory was kept off during script generation, so no run could benefit from
an earlier one. The comparison therefore isolates the contributions of the agentic
architecture and the scripting rules to script executability.

We developed a benchmark of eight prompts spanning common classes of MD calculation:
optimizing a lattice parameter by energy minimization, equilibrating a non-cubic crystal
under constant pressure, melting, equilibrating a slab with free surfaces, computing a
vacancy formation energy, measuring thermal expansion, uniaxial tensile loading, and
crack propagation under cyclic load. Each prompt fixes the material, the system size,
and the simulation protocol, and each names the OpenKIM model to be used. Naming the
model keeps potential selection out of the comparison, so that what is measured is the
construction of the script alone, and a failure can be traced to the script rather than
to an unsuitable or non-existent potential. Each prompt was issued three times to each
of six frontier models, Claude Opus 4.6, 4.8 and 5 and GPT-5.4, 5.5 and 5.6 Sol, under each
of the three conditions, giving 432 scripts in all. A script counts as successful only
if it runs to completion in LAMMPS. The prompts are listed in Section~S2 of the
supplementary information.

Figure~\ref{fig2} shows the outcome. Zero-shot prompting completes 92 of 144 scripts.
Routing the same request through ChatAtom raises this to 123, an improvement for every model
tested, and adding the scripting rules raises it again to 132.

\begin{figure}[h]
\centering
\includegraphics[width=0.85\textwidth]{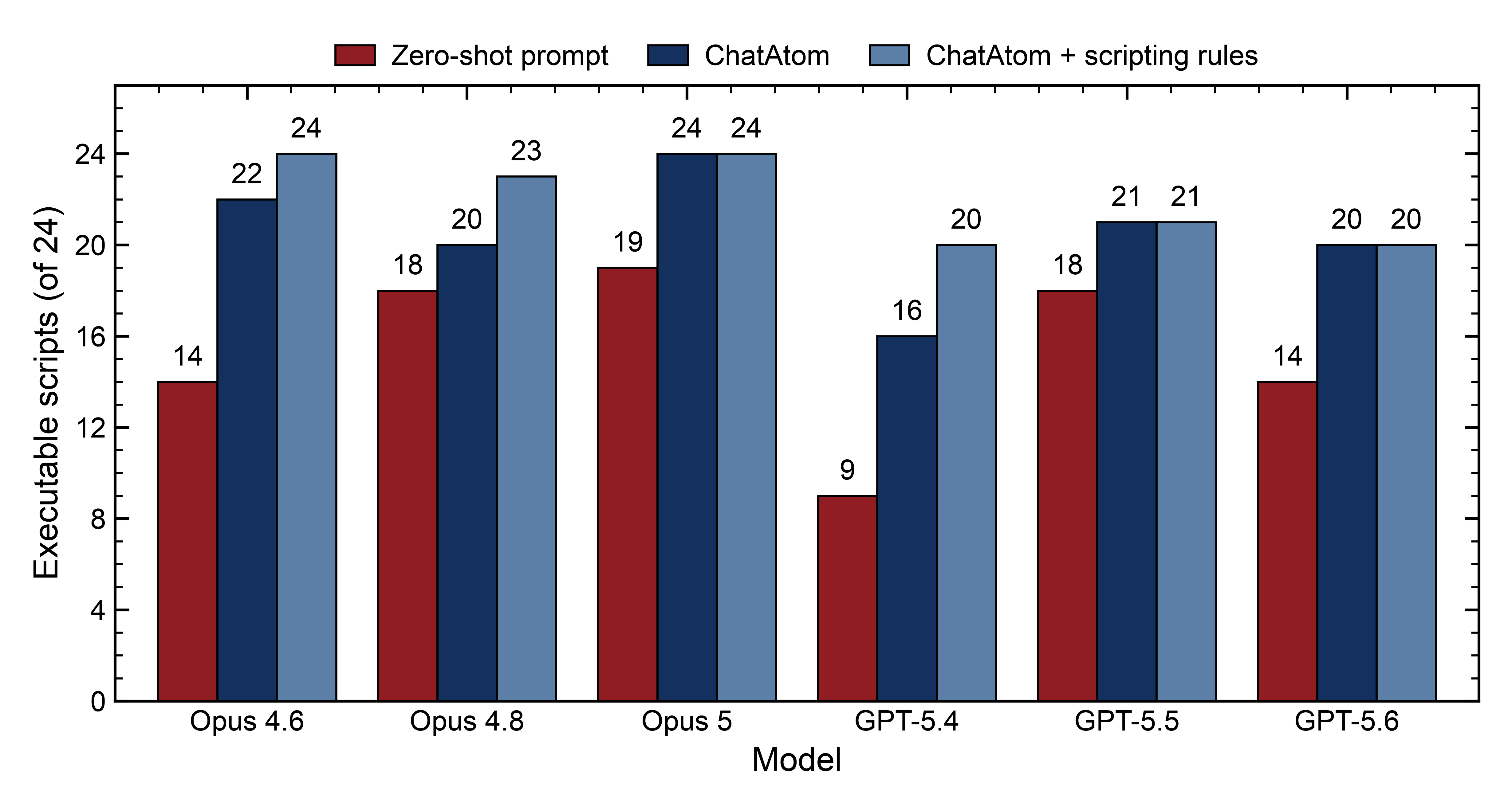}
\caption{Executable scripts produced under three conditions for six frontier models: zero-shot
prompting, ChatAtom, and ChatAtom with the LAMMPS scripting rules added to the knowledge base.
Each bar counts the scripts that ran to completion out of twenty-four, being eight
benchmark prompts issued three times. Pooled over all models the three conditions complete 92, 123 and 132 of 144 respectively.}\label{fig2}
\end{figure}

The failures are more informative than the totals. Every generated script was executed in
LAMMPS, and under zero-shot prompting fifty-two of them did not complete. Common issues
included incorrect ordering of the required KIM commands, such as placing \texttt{kim
interactions} before \texttt{kim init} or defining the simulation box first; use of a
thermodynamic keyword inside a variable without first including it in
\texttt{thermo\_style}; and use of lattice keywords that were incompatible with the
requested crystal structure, such as requesting a $c/a$ ratio from a lattice style for
which it is not defined. Other failures involved references to computes, fixes, or
variables that had been deleted or never defined, or the use of a keyword the command does not
recognize.

We then ran the same prompts through ChatAtom. Not one script broke the KIM command
ordering. That requirement is stated explicitly in the knowledge base of
Section~\ref{sec:lammps} and the reviewer is instructed to reject any script that violates
it. Twenty-one scripts still failed: some used a lattice keyword the requested crystal
does not have, and some referenced a fix or compute that had been deleted or never
defined.

We collected those remaining errors into a set of scripting rules and added them to the
knowledge base; the rules are reproduced in full in Section~S3 of the supplementary
information. Running the same prompts again with the rules in place raises the overall success rate across all models
from 123 to 132 of 144, and GPT-5.4 gains most, rising from 16 to 20. The model is
unchanged throughout, so the improvement comes from the knowledge base.

\begin{figure}[h]
\centering
\includegraphics[width=0.9\textwidth]{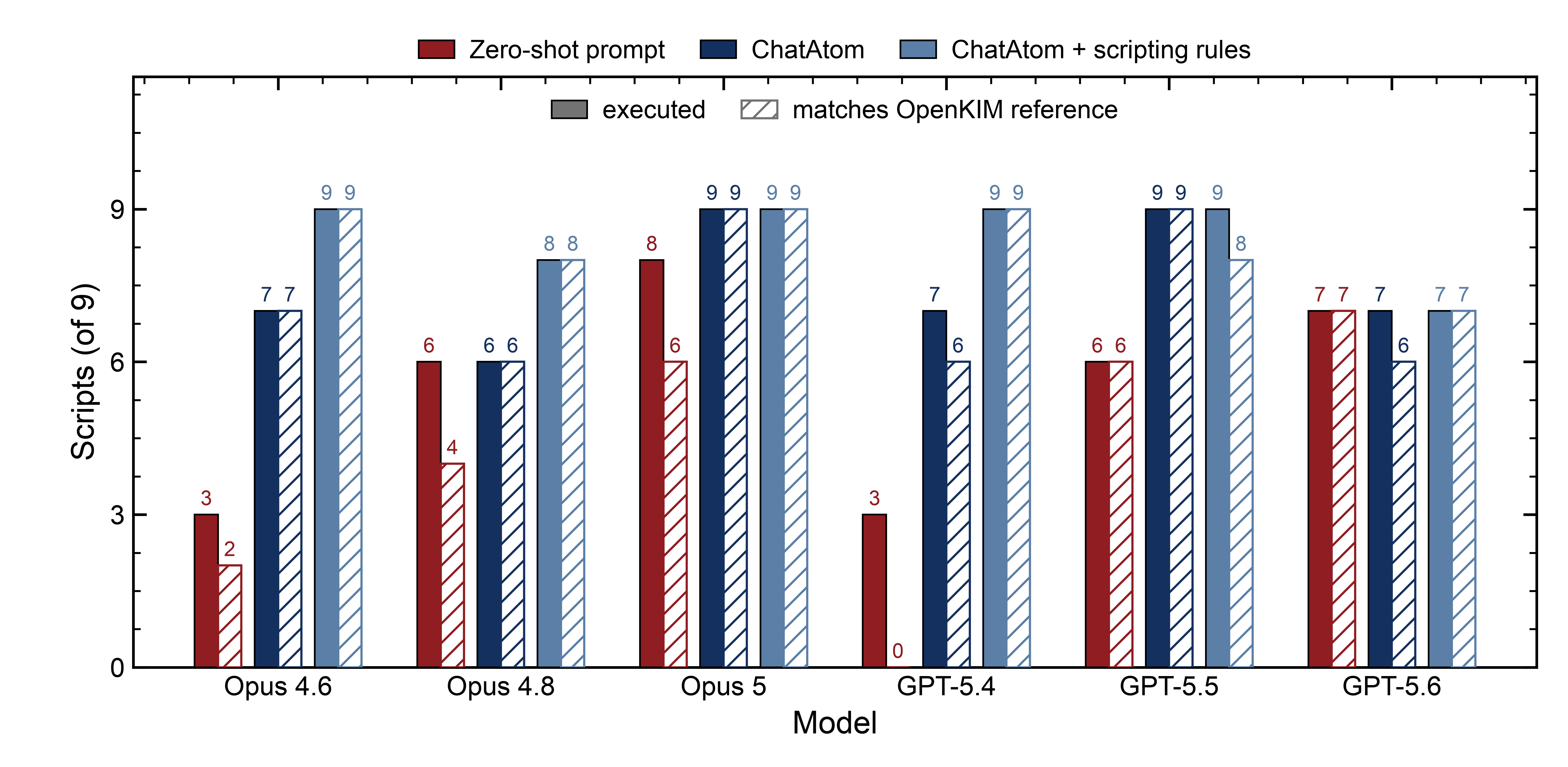}
\caption{Execution and physical correctness for the three benchmark prompts whose
quantity OpenKIM reports for the same interatomic potential: the lattice parameter of
fcc aluminum (P1), the vacancy formation energy of fcc copper (P5) and the thermal expansion
coefficient of bcc iron (P6), numbered as in Section~S2. Each model received the three prompts three times under each
condition, so nine scripts per bar. Solid bars count the scripts that ran to completion;
hatched bars count those whose returned value matches the OpenKIM reference. Pooled over
all models, 33, 45 and 51 of 54 scripts are executable and 25, 43 and 50 are physically
correct, for zero-shot prompting, ChatAtom, and ChatAtom with the scripting rules,
respectively.}\label{fig3}
\end{figure}

A script can run to completion and still return an incorrect physical result, so we also
evaluated the quantities computed by the generated scripts. Three of the eight benchmark
prompts produce quantities for which OpenKIM reports directly comparable reference
values for the same interatomic potential: the equilibrium lattice parameter of fcc
aluminum, 4.05~\AA; the vacancy formation energy of fcc copper, 1.27225~eV; and the
linear thermal expansion coefficient of bcc iron at 293~K,
$9.5673\times10^{-6}$~K$^{-1}$. The values returned by every generated script are listed
in Section~S4 of the supplementary information. For the lattice parameter, 10, 16, and
18 of 18 scripts executed under zero-shot prompting, ChatAtom, and ChatAtom with the
scripting rules, respectively, and every executable script reproduced the OpenKIM value
to within the 1\% threshold. For this calculation, execution success therefore
coincided with physical correctness. Figure~\ref{fig3} shows both measures for every model.

For the vacancy formation energy, 15, 15, and
16 of 18 scripts executed under zero-shot prompting, ChatAtom, and ChatAtom with the
scripting rules, respectively, but the number returning the
correct energy, within the 1\% threshold, was 8 of 15, 14 of 15, and 16 of 16.
The calculation needs the potential energy before and after the vacancy
is created. In each failing case the perfect-crystal energy was stored with a command of
the form \texttt{variable e0 equal pe}, which records the expression rather than its
value. When \texttt{e0} was read after the atom was removed, LAMMPS recomputed
\texttt{pe} for the already defective crystal, so both terms
in the expression referred to the same state, resulting in an incorrect vacancy energy.

The thermal expansion coefficient is obtained
from box lengths time-averaged at 253, 293 and 333~K, so a wider tolerance
is appropriate and a value is counted as correct within 5\% of the reference.
Out of 18 scripts per condition, 8 executed under zero-shot prompting,
14 within ChatAtom and 17 with the scripting rules; the number returning 
a correct value within the threshold was 7 of 8, 13 of 14 and 16 of 17. One zero-shot script, 
from Opus 4.6, returned exactly zero, failing like the vacancy scripts: 
the averaged box lengths were never saved, so all three were read at the 
end as the value from the last temperature. The remaining failure in each 
ChatAtom condition is of a different kind. Both let the sides of the box 
move independently instead of holding the cell cubic, so the box length
had not settled within the run length. The physics is correct; 
only the average is unconverged. Pooled over all models and the three prompts, 33, 45 and 51 of 54 scripts are executable
and 25, 43 and 50 return a value within the threshold, under zero-shot prompting, ChatAtom, and
ChatAtom with the scripting rules, respectively.

Taken together, these results indicate that whether a generated script successfully
executes, and whether it computes what was asked, depend not only on the underlying
model but also on the surrounding workflow and the knowledge available to it. Under
zero-shot prompting, the six models produced between 9 and 19 executable scripts out of
24; with the workflow and the scripting rules in place, this range narrowed to 20--24.
Model choice therefore remains important, but the differences between models in script
executability are reduced.

\subsection{End-to-end agentic workflow for crack propagation}

\label{sec:crack}
We next evaluated whether ChatAtom could construct and execute an end-to-end molecular
dynamics workflow from a high-level scientific request. As a case study, we chose crack
propagation, a representative materials science problem that proceeds in distinct
phases: the crystal is equilibrated at the target temperature, a crack is introduced,
the cracked configuration is relaxed, and mechanical loading is then applied. ChatAtom was
prompted to
\begin{quote}
\textit{``simulate crack propagation in copper at 300~K''}
\end{quote}
with the same
request subsequently applied to aluminum. No simulation parameters beyond the material,
temperature, and target phenomenon were specified. The objective was therefore not only
to generate an executable LAMMPS input script, but to evaluate whether ChatAtom could
translate the requested physical problem into a complete workflow encompassing
simulation design, script generation and review, execution, and post-simulation
analysis.

The orchestrator does not act on the request directly. Instead, it first returns a
complete proposal specifying the interatomic potential, boundary conditions, temperature, system size, crack
geometry, strain rate, frame-sampling frequency, and the ensemble and duration of each
simulation phase. No tool is called until the user approves the plan, which can be
revised over multiple turns as needed. This approval step converts an under-specified
request into an explicit simulation specification that can be inspected and modified
before computational resources are committed. Once the final plan was approved, 
ChatAtom proceeded to script generation and execution. The copper and
aluminum scripts were generated independently; Table~\ref{tab1} lists the configuration
for both metals, and the complete scripts are reproduced in Section~S5 of the
supplementary information.

The user approved ChatAtom's suggested crack geometry and the method of seeding. The seeding techniques available in the knowledge base include deleting atoms within a
region and excluding neighbor interactions across a plane.This choice
influences the simulation outcome and its interpretation. Deleting 200 atoms in a band
one lattice parameter wide removes two atomic layers over about 18~\AA{}, through the
full 3.6~nm periodic thickness of the cell.  Because the cell is also periodic in $x$, the notch repeats every 18~nm, and the
simulation loads a periodic array of interacting notches. Further, the seeded defect is therefore a blunt
nanoscale notch two atomic planes thick rather than an atomically sharp crack. A
defect this thin and this short sits between a nanocrack and a vacancy platelet, and
like a platelet it may collapse into a dislocation dipole or loop, because the two
free surfaces it creates carry an energy comparable to that of the collapsed
configuration. The 10~ps relaxation that follows seeding may be too short to reveal
such a collapse. The survival of the notch to the start of loading therefore does not
establish that it is stable. If the notch were to collapse over a longer relaxation,
the loading phase would test a defect cluster rather than a crack.  

\begin{table}[h]
\caption{Simulation settings for the copper and aluminum crack runs, taken from
the input scripts the LAMMPS agent produced.}\label{tab1}%
\footnotesize
\begin{tabular}{@{}lp{0.31\textwidth}p{0.31\textwidth}@{}}
\toprule
 & Copper & Aluminum \\
\midrule
OpenKIM model      & \texttt{EAM\_\allowbreak{}Dynamo\_\allowbreak{}Mishin\allowbreak{}Mehl\allowbreak{}Papaconstantopoulos\_\allowbreak{}2001\_\allowbreak{}Cu\_\_\allowbreak{}MO\_\allowbreak{}346334655118\_006} & \texttt{EAM\_\allowbreak{}Dynamo\_\allowbreak{}Mishin\allowbreak{}Farkas\allowbreak{}Mehl\_\allowbreak{}1999\_\allowbreak{}Al\_\_\allowbreak{}MO\_\allowbreak{}651801486679\_006} \\
Lattice parameter  & 3.615~\AA & 4.05~\AA \\
Atomic mass        & 63.546 & 26.9815 \\
CNA cutoff         & 3.086~\AA & 3.457~\AA \\
System size        & $50 \times 50 \times 10$ fcc unit cells ($10^{5}$ atoms) & $50 \times 50 \times 10$ fcc unit cells ($10^{5}$ atoms) \\
Crystal orientation & $x=[100]$, $y=[010]$, $z=[001]$ (LAMMPS default) & $x=[100]$, $y=[010]$, $z=[001]$ (LAMMPS default) \\
Boundary conditions & periodic in $x$, $y$ and $z$ & periodic in $x$, $y$ and $z$ \\
Temperature        & 300~K & 300~K \\
Timestep           & 1~fs & 1~fs \\

Phase 1            & NPT, 10~ps, 1~bar in $x$, $y$ and $z$ & NPT, 10~ps, 1~bar in $x$, $y$ and $z$ \\
Phase 2            & crack seeding (\texttt{delete\_atoms}) and NVT relaxation, 10~ps & crack seeding (\texttt{delete\_atoms}) and NVT relaxation, 10~ps \\
Phase 3            & NVT tension along $y$ at $10^{-3}$~ps$^{-1}$, 100~ps & NVT tension along $y$ at $10^{-3}$~ps$^{-1}$, 100~ps \\
Crack geometry     & 10\% of $L_x$, one lattice parameter wide (200 atoms removed) & 10\% of $L_x$, one lattice parameter wide (200 atoms removed) \\
Total trajectory   & 120~ps (10\% engineering strain) & 120~ps (10\% engineering strain) \\
\botrule
\end{tabular}
\end{table}

Once execution finished, the analysis agent was invoked to compute the stress--strain curve and the dislocation density.

\begin{figure}[h]
\centering
\includegraphics[width=0.9\textwidth]{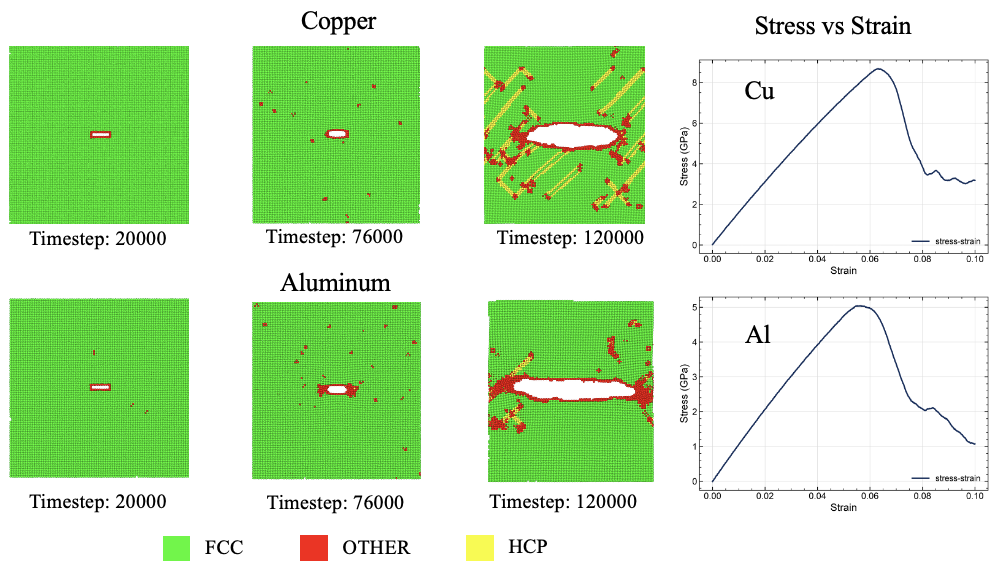}
\caption{Agent-generated crack simulations in copper (top) and aluminum (bottom) at
300~K, produced from a single natural-language request. Left to right: configurations at
the start of loading (step 20000), at 5.6\% strain (step 76000), and at 10\% strain
(step 120000), colored by common neighbor analysis (green, face-centered cubic; yellow,
hexagonal close packed, red, disordered; white, the crack
cavity), followed by the stress--strain response computed by the analysis
agent.}\label{fig4}
\end{figure}

Figure~\ref{fig4} shows trajectory frames at three timesteps alongside the
stress--strain response. At the start of loading both crystals contain the seeded crack.
At 5.6\% strain the crack has evolved into a rounded cavity while the surrounding
lattice remains almost entirely face-centered cubic. By 10\% strain copper has developed
an extended network of stacking faults reaching far from the crack, whereas in aluminum
the crack has extended along its original plane and the transformed material is confined
to the two crack tips.

The analysis agent computed the stress--strain curve for each run. For copper it reports
a nearly linear rise to about 8.6~GPa over the first 6.3\% of strain, a peak of 8.68~GPa
at $\varepsilon \approx 0.063$, then a sharp fall from 8.65 to 5.88~GPa within
$\Delta\varepsilon \approx 0.01$, after which the stress settles to 3.1--3.5~GPa from
$\varepsilon \approx 0.08$ and holds there to the end of loading. For aluminum it
reports a linear regime to $\varepsilon \approx 0.04$ and about 4~GPa, a peak of
5.04~GPa at $\varepsilon \approx 0.057$, a fall to roughly 2.3~GPa over the next 2\% of
strain, and a continued decline to about 1.1~GPa at 10\% strain. For copper it reports a
ductile fracture response, elastic loading to about 8.7~GPa followed by crack-tip
plasticity and blunting. 

For aluminum it reports: \textit{``The peak stress of $\approx 5.0$ GPa is high compared
to bulk experimental yield ($\approx 0.1$--0.3 GPa) because the system is nanoscale,
defect-free except for the seed crack, and loaded at an extreme strain rate
($10^{9}$~s$^{-1}$).''} It further reports that the sharp post-peak stress drop
indicates rapid crack propagation once the critical stress intensity is reached, while
the non-zero residual stress suggests the crack has not fully traversed the sample and
plastic deformation at the crack tip is absorbing energy.

These peak stresses exceed the strength of bulk copper and aluminum by more than an
order of magnitude, as is common in MD studies of strength and
fracture~\cite{Schiotz2003}. The simulated cell contains no dislocations before
loading, so plasticity must begin by nucleation, at stresses that approach the ideal
strength of the crystal~\cite{Ogata2002,Zhu2010}. The strain rate of about
$10^{9}$~s$^{-1}$ raises these stresses further, because it leaves little time for
thermally activated processes. The loading appears to hold the lateral
dimensions fixed, so much of the reported stress is hydrostatic rather than the shear
stress that drives plasticity. Defect-free specimens in experiment approach comparable
strengths~\cite{Richter2009}, but the values are not comparable to those of bulk
material. The contrast between the two metals, with copper emitting stacking faults
and aluminum extending its crack along the original plane, may also reflect the
crystal orientation and the stacking-fault and surface energies of each potential.
Taken together, the simulation describes void growth of a periodic array of blunt vacancy platelets
in an initially perfect crystal, loaded in uniaxial strain at about
$10^{9}$~s$^{-1}$. Its results describe that idealized system, not crack growth in a
bulk metal.  The analysis agent compared the peak stress with bulk yield for aluminum but missed it for
copper, and neither report mentions the remaining conditions, so a user without MD
experience could read the values as material properties.

Parts of the agent's interpretation go beyond the data. The first dislocation in
aluminum appears at about 5.6\% strain, at the peak stress, so the drop may reflect
dislocation nucleation as well as crack advance, and the stress--strain curve alone
cannot separate the two. The nonzero residual stress shows that an intact ligament still carries
load, which supports the agent's reading that the crack has not fully traversed the
cell. Dislocation activity has likely absorbed energy, but the residual stress does
not show it.

The analysis agent also ran a dislocation density analysis on each run. In copper the
first dislocations appear at step 88{,}000, six lines, all Shockley partials, at a
density of $8.1 \times 10^{15}$~m$^{-2}$. The density rises to a peak of
$2.95 \times 10^{17}$~m$^{-2}$ at
step 116{,}000, dominated by Shockley $\frac{1}{6}\langle 112 \rangle$ partials, which
carry 76\% of the total dislocation line length, followed by stair-rod
$\frac{1}{6}\langle 110 \rangle$ at 11\%. It falls by only 3\% to the final frame,
leaving a structure that is 82\% face-centered cubic, 9.5\% hexagonal close packed and
8.3\% other. The agent's reply for this run, as it appeared in the interface, is
reproduced in Figure~S4 of the supplementary information.

In aluminum the first dislocation is a single 17~\AA{} Shockley partial at step
76{,}000, and the density rises to a peak of $9.9 \times 10^{16}$~m$^{-2}$ at step
116{,}000, with Shockley
carrying 66\% of the line length and stair-rod 22\%. It then falls by 19\% to the final
frame, leaving 92.5\% face-centered cubic,
0.7\% hexagonal close packed and 6.8\% other. The peak density in copper is therefore
three times that in aluminum, consistent with copper blunting its crack by emitting
partials while aluminum extends along the original plane. The agent interprets both
metals as deforming by partial-mediated plasticity, and  the hexagonal close-packed fraction
as the stacking-fault ribbons.  Both interpretations appear to agree with Figure~\ref{fig4}.  The agent also interprets
the stair-rod content as Lomer--Cottrell
locks that impede further glide and contribute to strain hardening. This interpretation is plausible but overstated. In both metals the first dislocations appear at or after the peak stress, at about 6.8\% strain in copper and 5.6\% in aluminum. Any hardening from these dislocations would therefore appear after the peak. The stress–strain curves show no hardening after dislocation operation, although softening from growth of the cavity may mask it. Stair-rod segments may form locks, but the reported analysis classifies segments by Burgers vector and does not identify the lock configuration. The agent therefore stated a mechanism that is expected but not directly evidenced by the data.
For further detail, the analysis agent's
reports are reproduced verbatim in Section~S6 of the supplementary information.

\begin{figure}[h]
\centering
\includegraphics[width=0.9\textwidth]{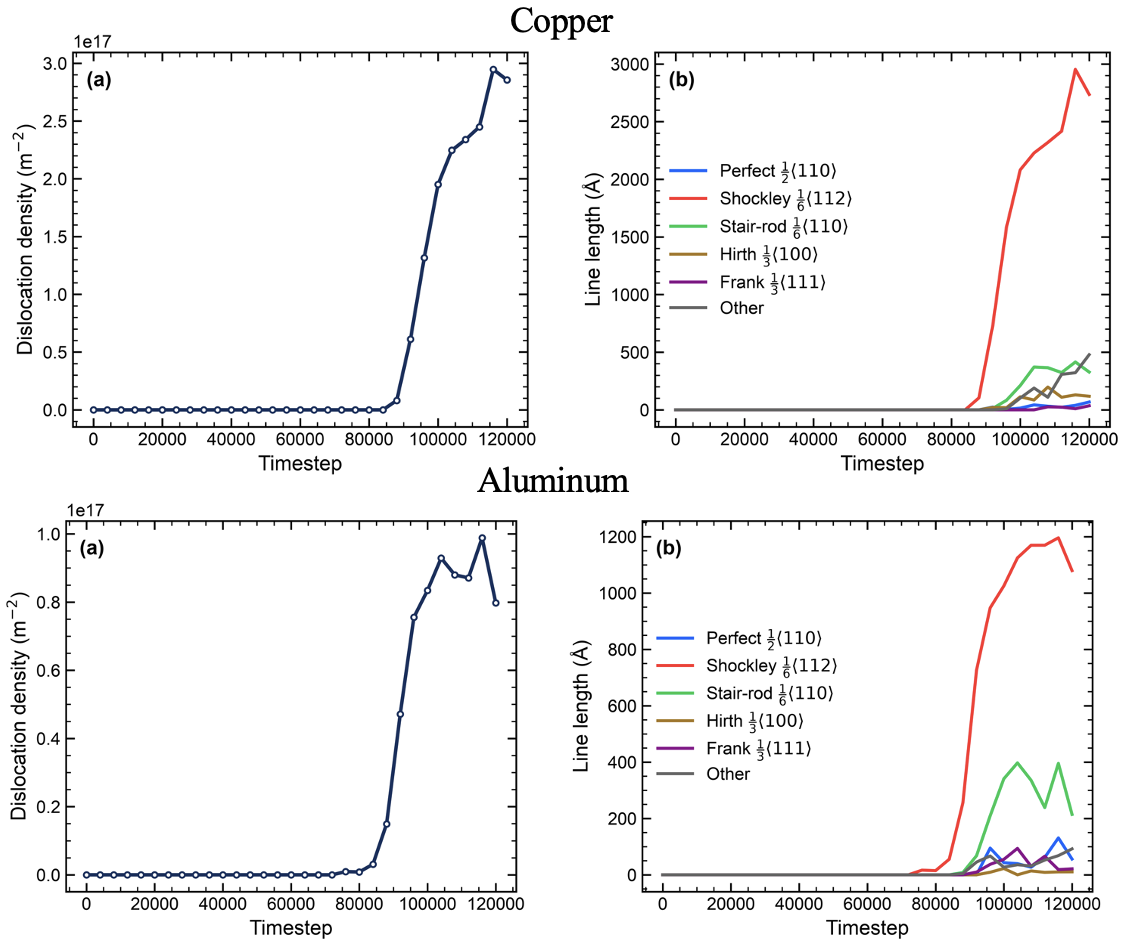}
\caption{Dislocation extraction analysis performed by the analysis agent on the
completed trajectories, for copper (top) and aluminum (bottom). (a)~Total dislocation
density against timestep. (b)~Line length resolved by Burgers-vector family. The agent
reports the first dislocations at step 88{,}000 in copper and at step 76{,}000 in
aluminum, peak densities of $2.95 \times 10^{17}$ and $9.9 \times 10^{16}$~m$^{-2}$, and the
Shockley $\frac{1}{6}\langle 112 \rangle$ family dominating the line content throughout
in both.}\label{fig5}
\end{figure}

\subsection{Autonomous hypothesis generation and testing}

\label{sec:campaign-results}

We next evaluated whether ChatAtom can generate its own hypotheses and test them
autonomously. The objective we set was to minimize the increase in the crack surface
area in copper during tensile loading. The user request was
\begin{quote}
\textit{``run a campaign on crack propagation in copper, minimizing the increase in the
crack surface area during tensile loading''}
\end{quote}
The orchestrator read this as a campaign
and loaded the skill described in Section~\ref{sec:campaign}, which sets out the procedure for running a campaign.
The orchestrator then returned a campaign proposal as defined in the skill: the system size, the interatomic
potential, the boundary conditions, the sequence of phases, the crack seeding technique, the parameters to sweep,
and the maximum number of iterations.

The parameters to sweep were the strain rate, from $10^{-3}$ to $2 \times 10^{-2}$~ps$^{-1}$,
the temperature, from 100 to 700~K, and the length of the initial crack, from 0.1 to 0.5
of the cell length. The LAMMPS agent wrote the script for the first trial in the usual
way, and that script became the template for the campaign: all other settings were fixed
in it and the user approved it once. The template is presented in Section~S7 of the supplementary
information. Every later trial was produced by substitution into that template rather
than by a fresh generation, so that a difference between trials could be attributed to
the parameters under test and not to a rewritten script. Each trial used a cell of
$60 \times 30 \times 3$ fcc unit cells, periodic in all directions. The crack was seeded by
excluding the interactions between the two groups of atoms facing each other across the
crack plane. The objective, the change in the crack surface area $\Delta a$, was measured by a tool
added for the campaign. It uses OVITO to build a surface mesh around the atoms, finds
the open surface at the crack, and measures its area at the end of relaxation and again
at the end of loading. The search works to keep the difference between the two as small
as it can, and a trial whose crack surface did not grow is considered to have a $\Delta a$ of 0. The loop was
given fifteen iterations and ran all fifteen successfully.

Before each trial the orchestrator recorded its reasoning, and that record shows how the
search proceeded. The first trial used mid-range values to establish a reference point,
and successive hypotheses then reduced the objective, first by lowering the strain rate
and the temperature, then by shortening the initial crack. The fifth trial, at
$10^{-3}$~ps$^{-1}$, 100~K and an initial crack length of 0.1 of the cell length,
returned no measurable crack growth and was the best of the campaign; the orchestrator
had predicted this outcome before running the trial, reasoning that a shorter initial crack
\textit{``should have a higher stress intensity factor threshold before propagation
begins, meaning less crack growth under the same loading''}. Rather than repeating the best trial, it spent the remaining trials probing
the surrounding parameter space, moving each parameter away from the best values to
locate the boundary between crack growth and no crack growth. Trials 6, 7 and 10 moved single parameters away from the best values to find
where the zero-growth regime ended, the sixth asking, \textit{``Now I want to find the
boundary, can we still get $\Delta$=0 with a larger crack? Testing crack\_frac=0.2
while keeping the optimal strain\_rate=0.001 and T=100 K.''} Trial 9 went to the opposite
extreme, \textit{``exploring the worst-case corner to understand the
full dynamic range: maximum strain rate (0.02), high temperature (700 K), and long crack
(0.5)''}, while trial 12 held the crack at its shortest under the same harsh loading,
testing whether \textit{``crack\_length\_frac=0.1 provides any protection under the
harshest conditions''}. None of those trials improved on the fifth. The full sequence of proposals,
parameters and measured objectives is given in Section~S8 of the supplementary
information.

The orchestrator's prediction for the fifth trial was correct, but its stated
reason misuses the concept of stress intensity factor.  The intended point is that a shorter
crack requires a higher applied stress before it propagates.  However, the stress intensity factor
threshold should be crack-length independent when properly applied. Moreover, stress intensity factor
commonly refers to the term from linear elastic fracture mechanics, which assumes a sharp 
crack whose elastic stress
field dominates a region large compared with the plastic process zone. An initial crack
six lattice parameters long, in a periodic cell three lattice parameters thick, does not
provide such a region. The agent therefore reached a correct conclusion but
used inaccurate and misleading terminology, a point we return to in
Section~\ref{sec:limitations}.

\begin{figure}[h]
\centering
\includegraphics[width=0.95\textwidth]{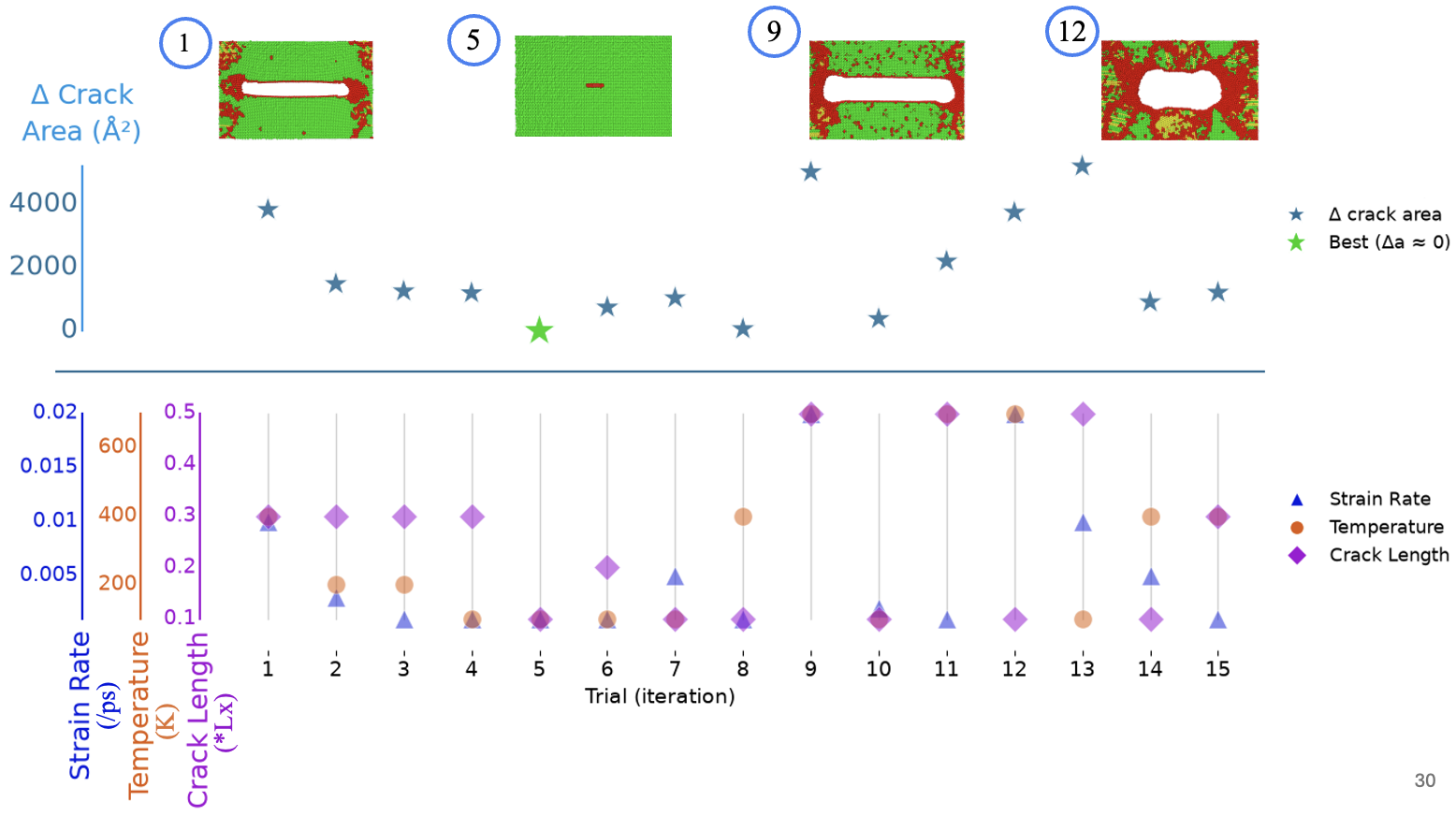}
\caption{Autonomous campaign over 15 iterations. Top: the measured change in
crack surface area for each trial, with the best trial marked, and the configuration at
the end of loading for trials 1, 5, 9 and 12. Bottom: the three parameter values
proposed by the orchestrator at each iteration. Trial 5, at the lowest strain rate, the
lowest temperature, and the shortest initial crack, showed no measured increase in
crack surface area, and the trials that followed explored the surrounding space without
improving on it.}\label{fig6}
\end{figure}

The report the agent generated at the end separates the effects of the three parameters
and is reproduced in Section~S9 of the supplementary information. It identifies the
strain rate as the dominant parameter and locates the threshold for zero propagation
sharply between $10^{-3}$ and $2 \times 10^{-3}$~ps$^{-1}$: the fifth trial, at the
lower rate, gave no growth, while the tenth, at twice the rate and otherwise identical,
gave 384~\AA$^2$. It places the initial crack length second, noting that lengthening the
crack from 0.1 to 0.2 and 0.3 of the cell at the same rate and temperature raised the
objective from zero to 745 and 1201~\AA$^2$. Temperature it ranks third: weak at low
rate and moderate crack length, where three trials differing only in temperature fall
between 1201 and 1258~\AA$^2$, but stronger at the shortest crack, where the fifth trial
at 100~K gave zero growth against 60~\AA$^2$ for the eighth at 400~K, which it interprets as
thermal fluctuations nucleating small voids. An increase of 60~\AA$^{2}$ corresponds to the area of a few surface atoms, and it
may instead reflect thermal roughening of the crack faces or error in the surface
mesh.  The agent further observes that at the highest
rate a short crack no longer helps: the twelfth trial reaching 3755~\AA$^2$ with the
shortest crack of the campaign. The ranking the agent suggests in this MD study rests on a small system, short timescales
and high strain rates, and may not hold for quasi-static crack propagation, where thermal
activation plays a larger role.

The campaign shows that ChatAtom autonomously proposes its own trials, each conditioned on
the results of those before it, and runs them to completion. It gained that capability
without any change to the agents or the execution path: the procedure entered as a
skill, and the operations it needed entered as campaign tools.

\section{Limitations and Future Directions}
\label{sec:limitations}

ChatAtom reduces the technical effort needed to conduct an MD study, but it does not
remove the need for scientific judgment. A simulation can satisfy a request while
adopting modeling choices that change its scientific meaning. In Section~3.2, the short,
periodically repeated defect grew into a rounded cavity in copper rather than behaving
as an isolated crack. Similarly, the autonomous campaign reached a correct prediction
but justified it using a stress intensity threshold that was inappropriate for the
simulated geometry. These examples distinguish successful execution and prediction
from sound physical interpretation.

Agent observability provides a way to make these limitations easier to identify and
address. Existing work considers complementary aspects of this problem: AgentOps
identifies the goals, plans, tools, and artifacts that should be traced throughout an
agent's lifecycle~\cite{dong2024agentops}, while AgentSight connects model interactions
to observed system activity~\cite{zheng2025agentsight}. Scientific observability
additionally requires explicit links between reported claims, supporting evidence,
and verification results~\cite{yin2026observability}. Before execution, ChatAtom could
compare the generated script with the approved plan and expose consequential choices,
including boundary conditions, loading constraints, and defect geometry. When a request
leaves these choices unresolved, the system should ask for clarification or explain
the interpretation it adopted. After execution, each conclusion should identify the
measurements that support it and distinguish direct observations from proposed
mechanisms. Plan revisions, user interventions, and unresolved disagreements should
remain linked to the run record. Such observability supports informed review without
itself guaranteeing that an interpretation is correct.

Scientific checks could strengthen this review through curated physical rules and
reference data. Rules should state not only a concept but also the conditions under
which it applies, while reference comparisons should account for cell size, strain
rate, and initial defect content. For example, the reported peak stresses and
dislocation densities should be contextualized rather than presented as transferable
bulk properties. Likewise, attributing hardening to Lomer--Cottrell locks requires
evidence beyond Burgers-vector populations; targeted analysis of dislocation-junction
topology could test that interpretation. Expert evaluation should assess the validity
of explanations and their evidential support separately from script executability or
prediction accuracy.

These extensions need not introduce a permanent agent for every additional
responsibility. Assumption checks, reference comparisons, and claim verification could
be supplied as loadable skills and deterministic tools used by the existing
orchestrator and analysis agent. Where additional review is warranted, a harness that
supports delegation could spawn a temporary, task-specific agent to examine modeling
assumptions or challenge an interpretation, with its findings retained in the same
observable record. More broadly, selected ChatAtom capabilities, including script
generation and review, simulation execution, visualization, and analysis, could be
packaged as plugins, reusable skills, and MCP tools for existing agent harnesses,
following the separation of domain tools from the harness advocated in our deployment
strategy at the Advanced Photon Source~\cite{du2026strategies}. Providing these
capabilities through familiar interfaces such as Claude Desktop or the Codex desktop
app could further reduce onboarding effort, provided that scientific checks, execution
approvals, and provenance are preserved.

The autonomous campaign also leaves opportunities to broaden the search strategy.
The current workflow selects trials sequentially through the orchestrator's reasoning
over previous results. Established optimizers, such as Bayesian optimization, could
be exposed through tools and skills so that the orchestrator can use a formal
optimizer, its own proposals, or a combination of the two. Batch proposals and
concurrent simulations could support searches that do not require every trial to
depend on the immediately preceding result.

\section{Conclusion}

In this work we presented ChatAtom, a multi-agent LLM system that turns a natural-language
request into a complete, reproducible molecular dynamics workflow without the user
writing a line of simulation code. A tool-calling orchestrator interprets each request,
selects an interatomic potential, and designs a multistage plan, then coordinates
specialized agents that write and review LAMMPS scripts, build OVITO visualization
pipelines, dispatch jobs to local or high-performance computing resources, and interpret
the resulting trajectories. Across eight benchmark tasks and six models, more scripts
ran to completion within ChatAtom than under zero-shot prompting of the same models, and
more of the values they returned were physically correct, within the threshold of the
OpenKIM reference. A single request was then expanded
into an executable, multi-phase crack simulation in copper and aluminum. The same
architecture closed into a campaign, supplied as a loadable skill, in which the system
autonomously generated its own hypotheses and tested them. We hope that ChatAtom provides
a foundation for agent-driven atomistic simulation, and that this integrated approach
lowers the barrier to MD studies and broadens access to advanced simulation workflows
for a wider community of researchers.

\backmatter

\bmhead{Supplementary information}

Supplementary information accompanies this manuscript. It contains the ChatAtom interface
during a run (S1), the benchmark prompt set (S2), the LAMMPS scripting rules supplied to
the generator and reviewer (S3), the value every generated script returned for the three
prompts with reference values (S4), the generated input scripts for both crack
simulations (S5), the analysis agent's reports in full (S6), the campaign template and
visualization scripts (S7), the per-iteration record of the autonomous campaign (S8),
and the report the agent produced at the end of it (S9).

\bmhead{Acknowledgements}

This work is supported by the U.S. Department of Energy, Office of Science, Office of
Advanced Scientific Computing Research and Office of Basic Energy Sciences, Scientific
Discovery through Advanced Computing (SciDAC) program under the MIRAGE project, award DE-SCL0000066. This research used resources of the National Energy Research Scientific Computing
Center (NERSC), a Department of Energy User Facility using NERSC award BES-
ERCAP0035465. We gratefully acknowledge the computing resources provided on Swing, a high-performance computing cluster operated by the Laboratory Computing Resource Center at Argonne National Laboratory. We thank Ryan Chard for helpful discussions and guidance on using Globus in ChatAtom. 

\mjc{The work of M.D. and M.J.C. at the Advanced Photon Source was also supported by the U.S. Department of Energy, Office of Science, Basic Energy Sciences, under Contract No.\ DE-AC02-06CH11357.}
 
\section*{Declarations}

\begin{itemize}
\item \textbf{Funding}: All the authors in this work are supported by the U.S\@.~Department of Energy, Office of Science, Office of Advanced Scientific Computing Research and  Office of Basic Energy Sciences, Scientific Discovery through Advanced Computing (SciDAC) program under the MIRAGE project, award DE-SCL0000066.
This research used resources of the National Energy Research Scientific Computing Center (NERSC), a Department of Energy User Facility using NERSC award BES-ERCAP0035465.
The initial template of atomistic simulation was developed at the Center for Integrated Nanotechnologies (CINT), an Office of Science user facility operated for the U.S\@.~Department of Energy.
This article has been authored by an employee of National Technology \& Engineering Solutions of Sandia, LLC under Contract No\@.~DE-NA0003525 with the U.S\@.~Department of Energy (DOE).
The employee owns all right, title, and interest in and to the article and is solely responsible for its contents. The United States Government retains and the publisher, by accepting the article for publication, acknowledges that the United States Government retains a non-exclusive, paid-up, irrevocable, world-wide license to publish or reproduce the published form of this article or allow others to do so, for United States Government purposes.
The DOE will provide public access to these results of federally sponsored research in accordance with the DOE Public Access Plan {https://www.energy.gov/downloads/doepublic-access-plan.} 
\item \textbf{Conflict of interest}: The authors have no conflicts to disclose.
\item \textbf{Ethics approval and consent to participate}: Not applicable
\item \mjc{\textbf{Data availability}: The data supporting this study are available in the Supplementary Information and the ChatAtom repository at \url{https://github.com/mirage-scidac/chatatom}. The Supplementary Information includes benchmark prompts, representative simulation and visualization scripts, reference-value comparisons, and autonomous campaign records.}
\item \textbf{Code availability}: The source code for ChatAtom is available on GitHub at \url{https://github.com/mirage-scidac/chatatom}.
\end{itemize}


\bibliography{references}

@article{Ogata2002,
  author  = {Ogata, Shigenobu and Li, Ju and Yip, Sidney},
  title   = {Ideal Pure Shear Strength of Aluminum and Copper},
  journal = {Science},
  volume  = {298},
  number  = {5594},
  pages   = {807--811},
  year    = {2002},
  doi     = {10.1126/science.1076652}
}

@article{Zhu2010,
  author  = {Zhu, Ting and Li, Ju},
  title   = {Ultra-strength materials},
  journal = {Progress in Materials Science},
  volume  = {55},
  number  = {7},
  pages   = {710--757},
  year    = {2010},
  doi     = {10.1016/j.pmatsci.2010.05.003}
}

@article{Richter2009,
  author  = {Richter, Gunther and Hillerich, Karla and Gianola, Daniel S. and
             M{\"o}nig, Reiner and Kraft, Oliver and Volkert, Cynthia A.},
  title   = {Ultrahigh Strength Single Crystalline Nanowhiskers Grown by Physical
             Vapor Deposition},
  journal = {Nano Letters},
  volume  = {9},
  number  = {8},
  pages   = {3048--3052},
  year    = {2009},
  doi     = {10.1021/nl9015107}
}

@article{Schiotz2003,
  author  = {Schi{\o}tz, Jakob and Jacobsen, Karsten W.},
  title   = {A Maximum in the Strength of Nanocrystalline Copper},
  journal = {Science},
  volume  = {301},
  number  = {5638},
  pages   = {1357--1359},
  year    = {2003},
  doi     = {10.1126/science.1086636}
}

@article{lammps2022,
  title   = {{LAMMPS} -- a flexible simulation tool for particle-based materials modeling at the atomic, meso, and continuum scales},
  author  = {Thompson, Aidan P. and Aktulga, H. Metin and Berger, Richard and Bolintineanu, Dan S. and Brown, W. Michael and Crozier, Paul S. and in 't Veld, Pieter J. and Kohlmeyer, Axel and Moore, Stan G. and Nguyen, Trung Dac and Shan, Ray and Stevens, Mark J. and Tranchida, Julien and Trott, Christian and Plimpton, Steven J.},
  journal = {Computer Physics Communications},
  volume  = {271},
  pages   = {108171},
  year    = {2022},
  doi     = {10.1016/j.cpc.2021.108171}
}

@article{openkim,
  title   = {The potential of atomistic simulations and the {K}nowledgebase of {I}nteratomic {M}odels},
  author  = {Tadmor, Ellad B. and Elliott, Ryan S. and Sethna, James P. and Miller, Ronald E. and Becker, Chandler A.},
  journal = {JOM},
  volume  = {63},
  number  = {7},
  pages   = {17},
  year    = {2011},
  doi     = {10.1007/s11837-011-0102-6}
}

@article{vriza2026multiagentic,
  title   = {Multi-agentic {AI} framework for end-to-end atomistic simulations},
  author  = {Vriza, Aikaterini and Kornu, Uma and Koneru, Aditya and Chan, Henry and Sankaranarayanan, Subramanian K. R. S.},
  journal = {Digital Discovery},
  volume  = {5},
  number  = {1},
  pages   = {440--452},
  year    = {2026},
  doi     = {10.1039/d5dd00435g}
}

@article{holbrook2026evaluating,
  title   = {Evaluating {LLM}-generated code for domain-specific languages: molecular dynamics with {LAMMPS}},
  author  = {Holbrook, Ethan and Verduzco, Juan C. and Strachan, Alejandro},
  journal = {Computational Materials Science},
  year    = {2026},
  note    = {arXiv:2603.20630},
  eprint  = {2603.20630},
  archive = {arXiv},
  archivePrefix = {arXiv}
}

@article{ko2026elagente,
  title  = {El {A}gente {P}otente: High-Throughput Agentic Atomistic Simulations},
  author = {Ko, Tsz Wai and Bai, Jiaru and Swanick, Thomas and Kang, Yeonghun and Choi, Changhyeok and Jiang, Angelina Qihong and Yin, Aiwei and Bernales, Varinia and Aspuru-Guzik, Al{\'a}n},
  year   = {2026},
  eprint = {2609.14840},
  archive = {arXiv},
  archivePrefix = {arXiv},
  primaryClass  = {cs.AI}
}

@article{wang2026agentmd,
  title  = {Agent-{MD}: Selective {LLM} Intervention with Event-Driven Escalation for Stateful {GCMC}--{MD} Campaigns},
  author = {Wang, Yijie and Yin, Zhen-Yu and Tang, Zhenheng and Chu, Xiaowen},
  year   = {2026},
  eprint = {2608.07637},
  archive = {arXiv},
  archivePrefix = {arXiv},
  primaryClass  = {cs.AI}
}

@article{shi2026mdagent2,
  title  = {{MDAgent2}: Large Language Model for Code Generation and Knowledge {Q}\&{A} in Molecular Dynamics},
  author = {Shi, Zhuofan and A, Hubao and Shao, Yufei and Huang, Dongliang and An, Hongxu and Xin, Chunxiao and Shen, Haiyang and Wang, Zhenyu and Na, Yunshan and Huang, Gang and Jing, Xiang},
  year   = {2026},
  eprint = {2601.02075},
  archive = {arXiv},
  archivePrefix = {arXiv},
  primaryClass  = {cs.CE}
}

@inproceedings{wang2026mdforge,
  title     = {{MDForge}: Agentic Molecular Dynamics Pipeline Design under Sparse Simulator Feedback},
  author    = {Wang, Zehong and Ma, Yijun and Schmidt, Connor R. and Ma, Tianyi and Sun, Weixiang and Li, Ziming and Guo, Xiaoguang and Zhang, Chuxu and Webber, Matthew J. and Ye, Yanfang},
  booktitle = {Proceedings of EMNLP 2026 (Main Conference)},
  year      = {2026},
  eprint    = {2606.12916},
  archive = {arXiv},
  archivePrefix = {arXiv}
}

@article{ma2026mdagent,
  title  = {{MDAgent}: A Multi-Agent Framework for End-to-End Molecular Dynamics Research},
  author = {Ma, Zhenyu and Yang, Chunyi and Song, Yuyang and Zhu, Jingyi and Yang, Letian and Xu, Limei and Xiao, Min and Jiang, Xukai},
  year   = {2026},
  eprint = {2604.18622},
  archive = {arXiv},
  archivePrefix = {arXiv},
  primaryClass  = {q-bio.QM}
}

@article{dspy,
  title   = {{DSPy}: Compiling Declarative Language Model Calls into Self-Improving Pipelines},
  author  = {Khattab, Omar and Singhvi, Arnav and Maheshwari, Paridhi and Zhang, Zhiyuan
             and Santhanam, Keshav and Vardhamanan, Sri and Haq, Saiful and Sharma, Ashutosh
             and Joshi, Thomas T. and Moazam, Hanna and Miller, Heather and Zaharia, Matei
             and Potts, Christopher},
  journal = {arXiv preprint arXiv:2310.03714},
  year    = {2023},
  doi     = {10.48550/arXiv.2310.03714}
}

@article{somasundaram2026ursa,
  title   = {An Agentic Orchestration of Atomistic Simulations},
  author  = {Somasundaram, Rahul and Habib, Adela and Dang, Khanh and Shivakumar, Sachin
             and Hill, Ryley G. and Wimmer, Golo and Mishra, Avanish and Pachalieva, Aleksandra
             and Lui, Arthur and Viswanathan, Hari and Grosskopf, Michael and Fensin, Saryu
             and Bent, Russell and DeBardeleben, Nathan and Lawrence, Earl},
  journal = {arXiv preprint arXiv:2607.22596},
  year    = {2026}
}

@article{liu2026masgent,
  title   = {Masgent: an {AI}-assisted materials simulation agent},
  author  = {Liu, Guangchen and Yang, Songge and Zhong, Yu},
  journal = {Digital Discovery},
  volume  = {5},
  number  = {5},
  pages   = {2151--2171},
  year    = {2026},
  doi     = {10.1039/D6DD00043F}
}

@article{tuckerman2000understanding,
  title   = {Understanding Modern Molecular Dynamics: Techniques and Applications},
  author  = {Tuckerman, Mark E. and Martyna, Glenn J.},
  journal = {The Journal of Physical Chemistry B},
  volume  = {104},
  number  = {2},
  pages   = {159--178},
  year    = {2000},
  doi     = {10.1021/jp992433y}
}

@article{ovito,
  title   = {Visualization and analysis of atomistic simulation data with {OVITO} -- the Open Visualization Tool},
  author  = {Stukowski, Alexander},
  journal = {Modelling and Simulation in Materials Science and Engineering},
  volume  = {18},
  number  = {1},
  pages   = {015012},
  year    = {2010},
  doi     = {10.1088/0965-0393/18/1/015012}
}

@article{stukowski2012dxa,
  title   = {Automated identification and indexing of dislocations in crystal interfaces},
  author  = {Stukowski, Alexander and Bulatov, Vasily V. and Arsenlis, Athanasios},
  journal = {Modelling and Simulation in Materials Science and Engineering},
  volume  = {20},
  number  = {8},
  pages   = {085007},
  year    = {2012},
  doi     = {10.1088/0965-0393/20/8/085007}
}

@article{stukowski2012structure,
  title   = {Structure identification methods for atomistic simulations of crystalline materials},
  author  = {Stukowski, Alexander},
  journal = {Modelling and Simulation in Materials Science and Engineering},
  volume  = {20},
  number  = {4},
  pages   = {045021},
  year    = {2012},
  doi     = {10.1088/0965-0393/20/4/045021}
}

@article{boiko2023autonomous,
  title   = {Autonomous chemical research with large language models},
  author  = {Boiko, Daniil A. and MacKnight, Robert and Kline, Ben and Gomes, Gabe},
  journal = {Nature},
  volume  = {624},
  pages   = {570--578},
  year    = {2023},
  doi     = {10.1038/s41586-023-06792-0}
}

@article{mbran2024chemcrow,
  title   = {Augmenting large language models with chemistry tools},
  author  = {M. Bran, Andres and Cox, Sam and Schilter, Oliver and Baldassari, Carlo
             and White, Andrew D. and Schwaller, Philippe},
  journal = {Nature Machine Intelligence},
  volume  = {6},
  pages   = {525--535},
  year    = {2024},
  doi     = {10.1038/s42256-024-00832-8}
}

@article{wang2025dreams,
  title  = {{DREAMS}: Density Functional Theory Based Research Engine for Agentic Materials Simulation},
  author = {Wang, Ziqi and Huang, Hongshuo and Zhao, Hancheng and Xu, Changwen and Zhu, Shang and Janssen, Jan and Viswanathan, Venkatasubramanian},
  year   = {2025},
  eprint = {2507.14267},
  archive = {arXiv},
  archivePrefix = {arXiv},
  primaryClass  = {cs.AI}
}

@article{jacobsson2026hypotheses,
  title   = {{AI}-Generated Hypotheses and the Emergence of Autonomous Scientific Discovery},
  author  = {Jacobsson, T. Jesper},
  journal = {ACS Materials Letters},
  volume  = {8},
  pages   = {1457--1464},
  year    = {2026},
  doi     = {10.1021/acsmaterialslett.6c00224}
}

@inproceedings{kumbhar2025hypothesis,
  title     = {Hypothesis Generation for Materials Discovery and Design Using Goal-Driven and Constraint-Guided {LLM} Agents},
  author    = {Kumbhar, Shrinidhi and Mishra, Venkatesh and Coutinho, Kevin and Handa, Divij and Iquebal, Ashif and Baral, Chitta},
  booktitle = {Findings of the Association for Computational Linguistics: {NAACL} 2025},
  year      = {2025},
  doi       = {10.18653/v1/2025.findings-naacl.420}
}

@article{roy2026autonomous,
  title   = {Autonomous alloy composition optimization using molecular dynamics guided by a large language model},
  author  = {Roy, Ankit and Devanathan, Ram and Nartu, M. S. K. K. Y. and Joshi, Vineet V.},
  journal = {Scripta Materialia},
  volume  = {272},
  pages   = {117048},
  year    = {2026},
  doi     = {10.1016/j.scriptamat.2025.117048}
}

@article{zhou2026autonomous,
  title  = {Autonomous Agents for Scientific Discovery: Orchestrating Scientists, Language, Code, and Physics},
  author = {Zhou, Lianhao and Ling, Hongyi and Fu, Cong and Huang, Yepeng and Sun, Michael and
            Yu, Wendi and Wang, Xiaoxuan and Li, Xiner and Su, Xingyu and Zhang, Junkai and
            Chen, Xiusi and Liang, Chenxing and Qian, Xiaofeng and Ji, Heng and Wang, Wei and
            Zitnik, Marinka and Ji, Shuiwang},
  year   = {2026},
  eprint = {2510.09901},
  archive = {arXiv},
  archivePrefix = {arXiv},
  primaryClass  = {cs.AI}
}

@misc{langgraph,
  title        = {{LangGraph}: a low-level orchestration framework for building stateful agents},
  author       = {{LangChain, Inc.}},
  year         = {2024},
  howpublished = {\url{https://github.com/langchain-ai/langgraph}},
  note         = {Software; accessed 6 October 2026}
}

@misc{globusmcp,
  title        = {{Globus MCP} Server: an {MCP} server implementation enabling {AI} applications to interact with {Globus} services},
  author       = {{Globus}},
  howpublished = {\url{https://github.com/globus/globus-mcp}},
  note         = {Software; accessed 6 October 2026}
}

@article{dong2024agentops,
  title   = {{AgentOps}: Enabling Observability of {LLM} Agents},
  author  = {Dong, Liming and Lu, Qinghua and Zhu, Liming},
  journal = {arXiv preprint arXiv:2411.05285},
  year    = {2024},
  eprint  = {2411.05285},
  archivePrefix = {arXiv},
  primaryClass  = {cs.AI},
  doi     = {10.48550/arXiv.2411.05285},
  url     = {https://arxiv.org/abs/2411.05285}
}

@inproceedings{zheng2025agentsight,
  title     = {{AgentSight}: System-Level Observability for {AI} Agents Using {eBPF}},
  author    = {Zheng, Yusheng and Hu, Yanpeng and Yu, Tong and Quinn, Andi},
  booktitle = {Proceedings of the 4th Workshop on Practical Adoption Challenges of ML for Systems},
  series    = {PACMI '25},
  pages     = {110--115},
  year      = {2025},
  publisher = {ACM},
  doi       = {10.1145/3766882.3767169}
}

@article{yin2026observability,
  title   = {Artifact-centered Claim-aware Observability for Autonomous Scientific Agents},
  author  = {Yin, Xiangyu and Du, Ming and Prince, Michael H. and Cherukara, Mathew J.},
  journal = {arXiv preprint arXiv:2608.18312},
  year    = {2026},
  eprint  = {2608.18312},
  archivePrefix = {arXiv},
  primaryClass  = {cs.CL},
  doi     = {10.48550/arXiv.2608.18312},
  url     = {https://arxiv.org/abs/2608.18312}
}

@article{du2026strategies,
  title   = {Strategies for Deploying {AI} Agents in Production at Scientific User Facilities},
  author  = {Du, Ming and Yin, Xiangyu and Prince, Michael and Jiang, Yi and
             Sainju, Rajat and Bicer, Tekin and Luo, Yanqi and Codrea, Eric and
             Myint, Peco and Andrejevic, Nina and Huang, Juanjuan and Mohanty, Trupti and
             Tripathi, Pawan and Beniwal, Dishant and Sharma, Hemant and Gursoy, Doga and
             Luo, Aileen and Zhou, Tao and Xu, Chenran and Ilavsky, Jan and
             Dearing, Matthew T. and Chard, Ryan and Seo, Hoon and Jarosz, Dariusz and
             Chandler, Elaine and Fors, Thomas and Shang, Hairong and Jemian, Pete and
             Wu, Xuli and Miller, Madeline and Aydelott, Ryan and Schwarz, Nicholas and
             de Carlo, Francesco and Lee, Byeongdu and Sun, Yine and Smart, Moises and
             Piot, Philippe and Sullivan, Joe and Cherukara, Mathew J. and Sandy, Alec and
             Vogt, Stefan and Lang, Jonathan and Chapon, Laurent C.},
  journal = {arXiv preprint arXiv:2609.36362},
  year    = {2026},
  eprint  = {2609.36362},
  archivePrefix = {arXiv},
  primaryClass  = {cs.SE},
  doi     = {10.48550/arXiv.2609.36362},
  url     = {https://arxiv.org/abs/2609.36362}
}

@article{boyce2026materials,
  title   = {Materials Behavior as Mechanism Ensembles: A Probabilistic Framework for Emergent Behaviors},
  author  = {Boyce, Brad L. and Wood, Mitchell A. and Garikipati, Krishna and Robertson, Andreas E. and Larson, Jeffrey and Srivastava, Ishan and Debusschere, Bert and Desai, Saaketh and Iyer, Prasad and Robbe, Pieterjan and Cherukara, Mathew and Munson, Todd and Du, Ming and Mohanty, Trupti and Gardner, David J. and Capolungo, Laurent and Jasperson, Benjamin A. and Tan, Jingye and Dingreville, R{\'e}mi},
  year    = {2026},
  eprint  = {2607.27163},
  archivePrefix = {arXiv},
  primaryClass  = {cond-mat.mtrl-sci}
}

\end{document}


\rhead{}
\maketitle


\section{The ChatAtom interface}
\label{sup:implementation}

\begin{figure}[H]
  \centering
  \includegraphics[width=0.98\textwidth]{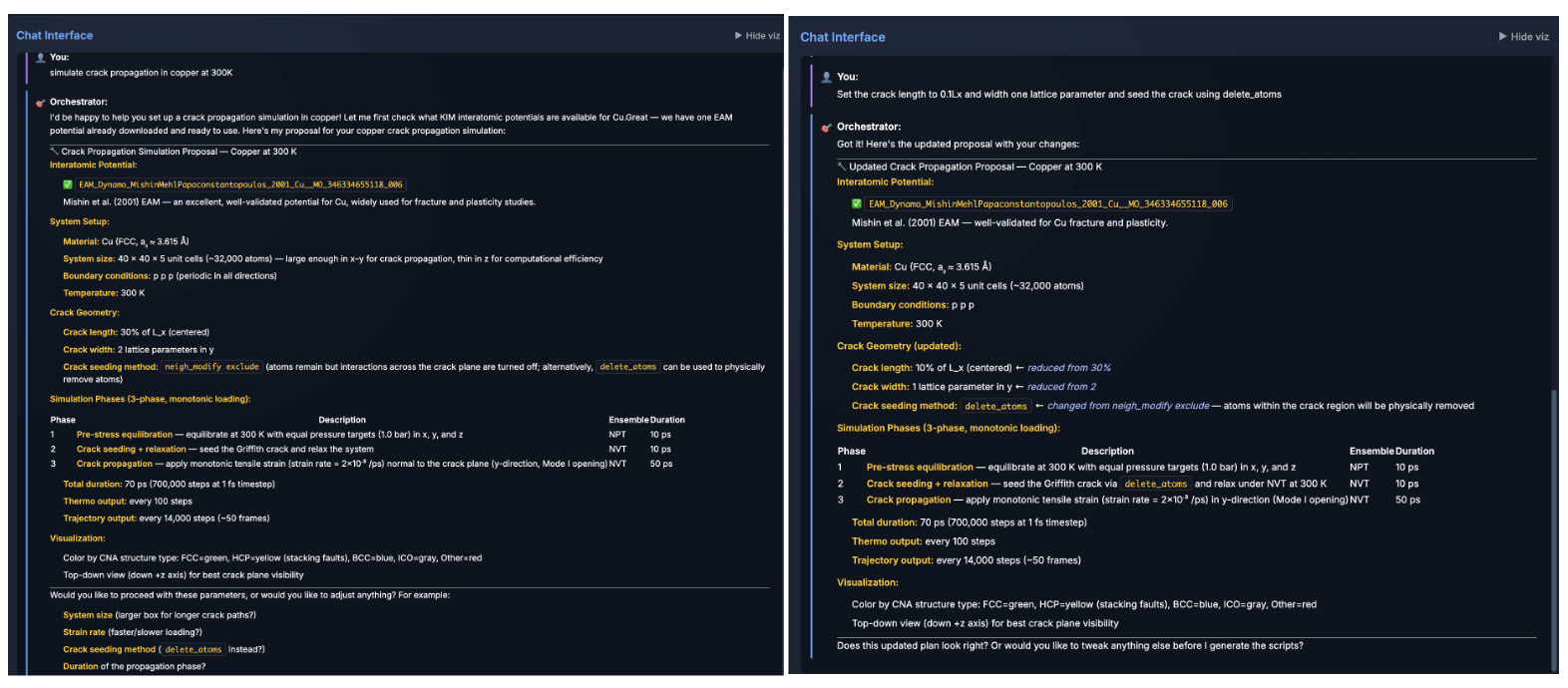}
  \caption{Multi-turn planning in the orchestrator. Left: the proposal returned for the request \textit{``simulate crack propagation in copper at 300 K''}. Right: the updated proposal after the user asked for a shorter crack, a narrower crack width and a different seeding method, with each revised value shown against the one it replaces.}
  \label{fig:sfigplan}
\end{figure}

Figure~\ref{fig:sfigplan} shows multiturn conversation between user and orchestrator. The user asks for a crack propagation simulation in copper, and the orchestrator answers with a
proposal naming the interatomic model, the system size, the three simulation phases with their ensembles and durations, the crack geometry and the visualization, then asks for approval. When the user changes the crack length, the crack width and the
seeding method, the orchestrator returns an updated proposal in which each changed value is shown against the one it replaces.

\begin{figure}[H]
  \centering
  \includegraphics[width=0.98\textwidth]{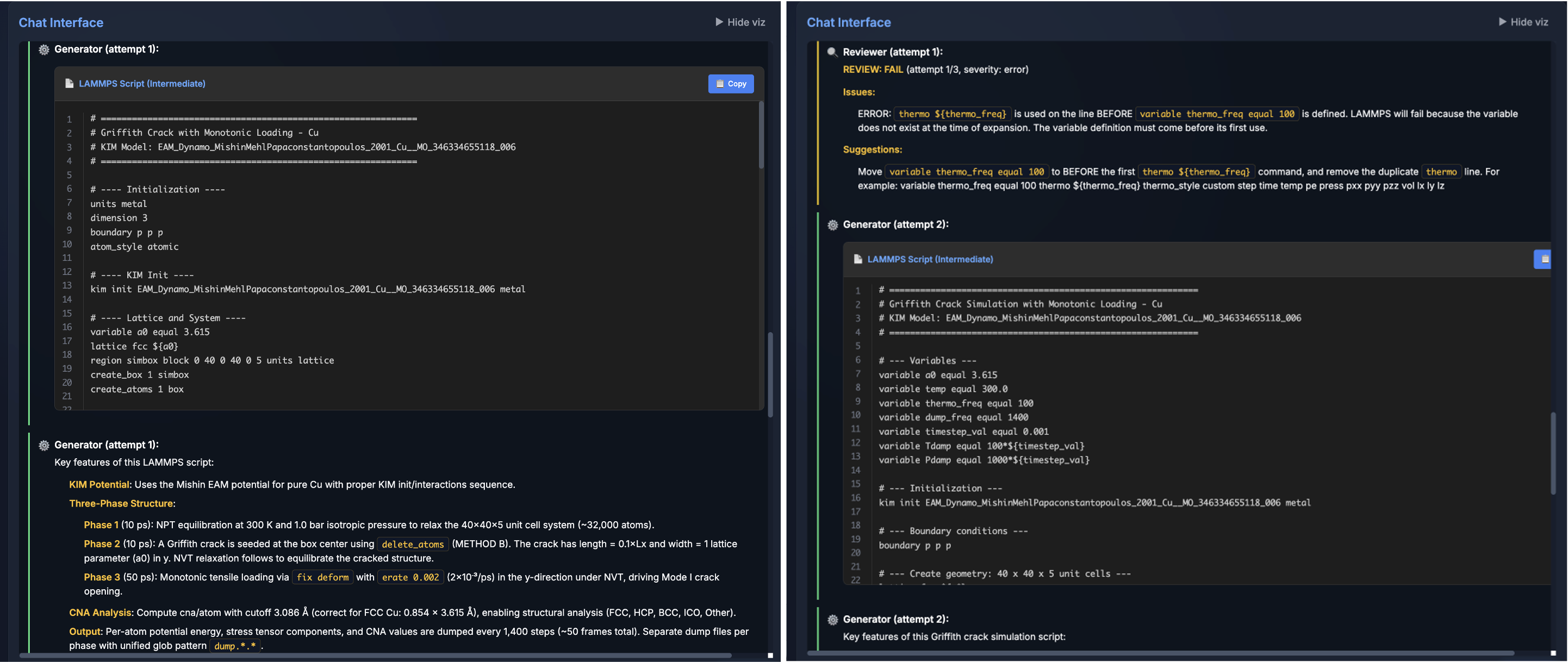}
  \caption{The generate--review loop of the LAMMPS agent. Left: the generator's first draft, with its own summary of the script below it. Right: the reviewer rejecting that draft because a thermodynamic keyword is used before the variable it refers to is defined, together with the fix it proposes, followed by the generator's second attempt, in which the variables are defined first.}
  \label{fig:sfiggenrev}
\end{figure}

Figure~\ref{fig:sfiggenrev} shows the generate--review loop of the LAMMPS agent in the same interface. The generator drafts a script and summarizes what it has written; the reviewer reads that draft, returns a verdict with the error it found and the fix it proposes, and the generator then writes a fresh script.

\begin{figure}[H]
  \centering
  \includegraphics[width=0.98\textwidth]{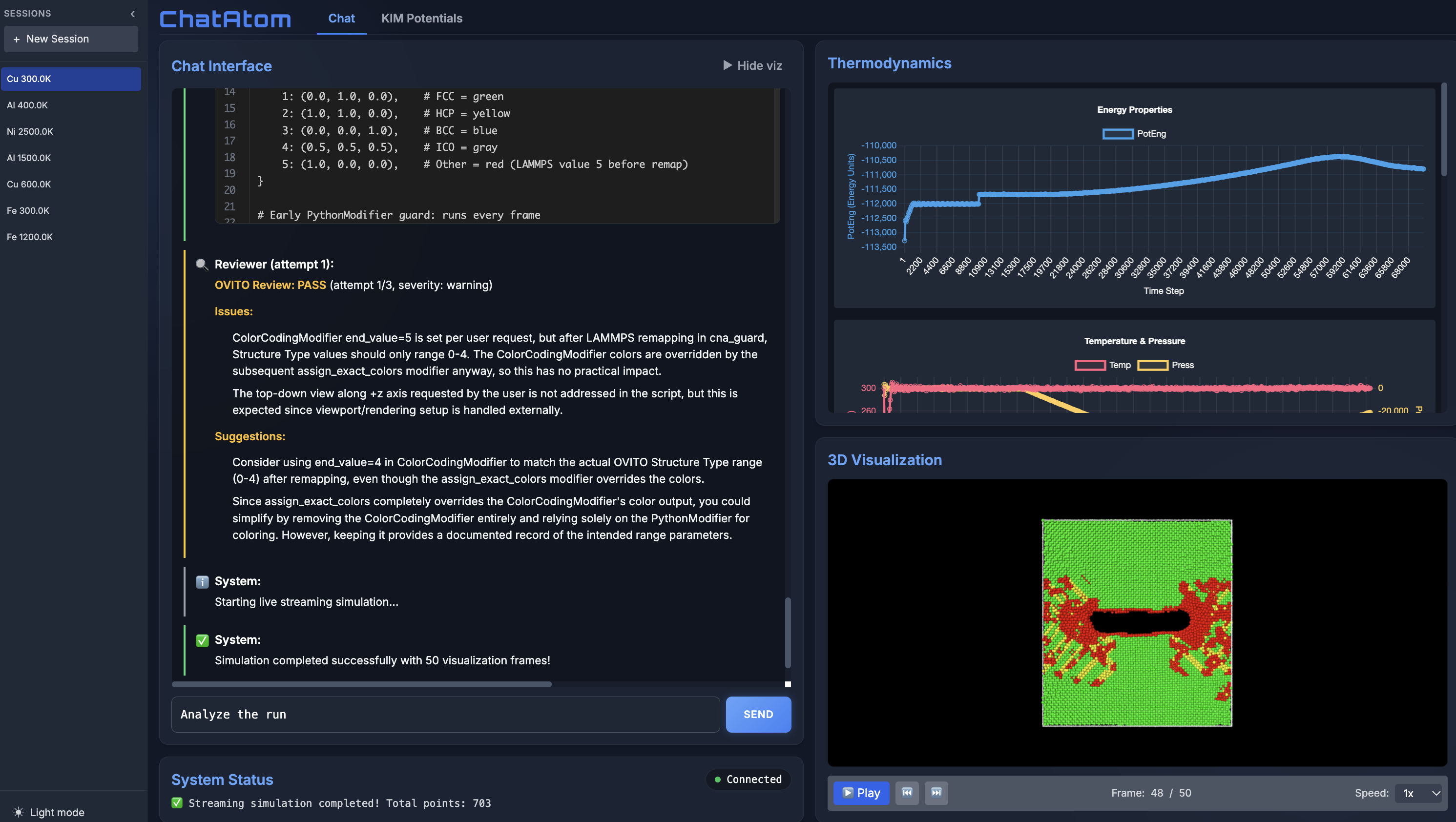}
  \caption{The ChatAtom web interface: session list (left), natural-language
  conversation with streaming agent progress (center), and thermodynamic
  plots and trajectory rendering (right).}
  \label{fig:sfig1}
\end{figure}

Figure~\ref{fig:sfig1} shows the ChatAtom user interface.

\begin{figure}[H]
  \centering
  \includegraphics[width=0.98\textwidth]{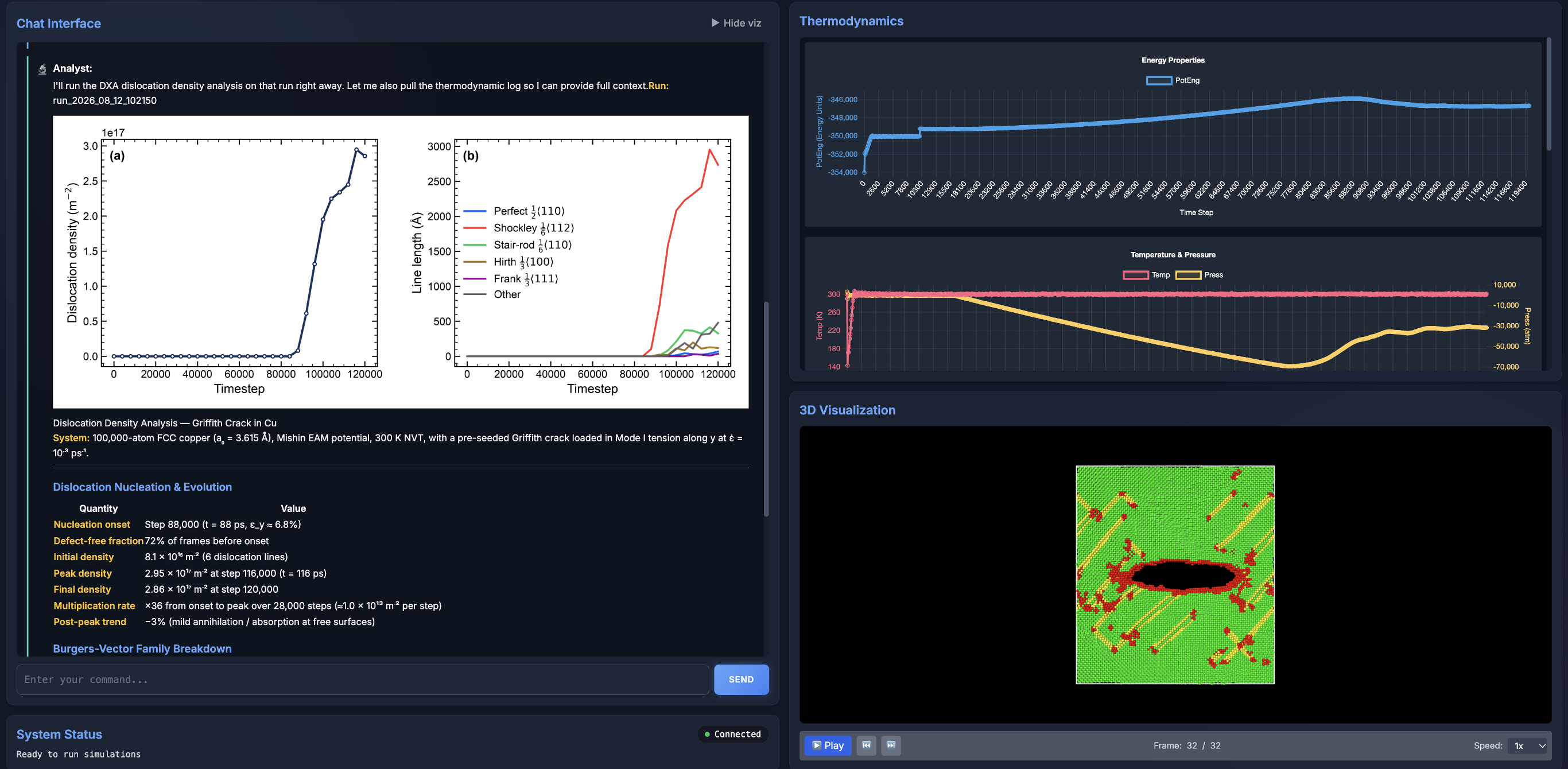}
  \caption{The analysis agent reporting in the interface. The chat pane holds the agent's reply to a request for a dislocation analysis of the copper run: the total dislocation density and the line length resolved by Burgers-vector family, followed by the written report and its table of nucleation onset, densities and multiplication rate. The thermodynamic panels and the rendered configuration on the right belong to the same run.}
  \label{fig:sfiganalysis}
\end{figure}

Figure~\ref{fig:sfiganalysis} shows an analysis returned in the interface. The user asked for a dislocation analysis of the copper run reported in the main text, and the analysis agent ran dislocation extraction on the stored trajectory and replied in the chat pane with the plots and a written report, while the panels on the right continued to show the thermodynamic output and the rendered configuration for the same run.

\section{Benchmark prompt set}
\label{sup:prompts-benchmark}

The prompts below define the benchmark used to compare ChatAtom against frontier models
prompted zero-shot. Each prompt fixes the material, the OpenKIM interatomic model, the
system size, and the ensemble sequence.
Each prompt was issued three times to every model, giving three independent replicates per
prompt and model.

\begin{enumerate}[label=\textbf{P\arabic*.},leftmargin=2.4em]

\item \textit{Energy minimization lattice param}\\
Simulate FCC aluminum to determine its optimized lattice parameter. Use the KIM potential EAM\_\allowbreak{}Dynamo\_\allowbreak{}MishinFarkasMehl\_\allowbreak{}1999\_\allowbreak{}Al\_\allowbreak{}\_\allowbreak{}MO\_\allowbreak{}651801486679\_\allowbreak{}006 . Build a 4 $\times$ 4 $\times$ 4 unit cell system, periodic in all directions, starting from a lattice parameter of 4 Angstrom. Relax both the atomic positions and the simulation-cell dimensions to zero pressure using conjugate-gradient (cg) energy minimization, with energy and force tolerances of 1 $\times$ 10\textasciicircum{}-8 and limits of 10000 iterations and 100000 force evaluations. At the end, calculate the optimized lattice constant from the relaxed cell dimensions and report it, along with the number of atoms and the final pressure.

\item \textit{NPT  Equilibrium}\\
Simulate the equilibration of bulk titanium at 300 K and 1 bar. Use the KIM potential EAM\_\allowbreak{}Dynamo\_\allowbreak{}Ackland\_\allowbreak{}1992\_\allowbreak{}Ti\_\allowbreak{}\_\allowbreak{}MO\_\allowbreak{}748534961139\_\allowbreak{}006. Build titanium in its equilibrium crystal structure with an 8x8x6 unit cell supercell, periodic in all directions. Initialise velocities at 300 K, then run in the NPT ensemble at 300 K and 1 bar for 30 ps. Print the three box lengths in the thermodynamic output every 100 steps, calculate and print the relaxed c/a ratio and the number of atoms.

\item \textit{Melting}\\
Simulate melting of FCC nickel using the OpenKIM potential EAM\_\allowbreak{}Dynamo\_\allowbreak{}AcklandTichyVitek\_\allowbreak{}1987\_\allowbreak{}Ni\_\allowbreak{}\_\allowbreak{}MO\_\allowbreak{}977363131043\_\allowbreak{}006. Build an 8 $\times$ 8 $\times$ 8 unit cell system with periodic boundary conditions. First equilibrate the crystal at 300 K and 1 bar for 10 ps in the NPT ensemble. Then heat it continuously from 300 K to 2500 K over 40 ps at 1 bar, followed by a 20 ps hold at 2500 K. Print the timestep, temperature, pressure, potential energy, kinetic energy, total energy, volume, box dimensions, and density every 100 timesteps.

\item \textit{NVT Equilibrium}\\
Simulate the equilibration of an aluminum slab at 300 K. Use the KIM potential EAM\_\allowbreak{}Dynamo\_\allowbreak{}ErcolessiAdams\_\allowbreak{}1994\_\allowbreak{}Al\_\allowbreak{}\_\allowbreak{}MO\_\allowbreak{}123629422045\_\allowbreak{}005. Build an FCC aluminum slab of 8x8x12  unit cell system, periodic in x and y and non-periodic along z. Initialise velocities at 300 K, then run in the NVT ensemble at 300 K for 10 ps. Print the potential energy in the thermodynamic output every 100 steps.

\item \textit{Vacancy}\\
Calculate the vacancy formation energy of FCC copper using LAMMPS. Use the KIM potential EAM\_\allowbreak{}Dynamo\_\allowbreak{}MishinMehlPapaconstantopoulos\_\allowbreak{}2001\_\allowbreak{}Cu\_\allowbreak{}\_\allowbreak{}MO\_\allowbreak{}346334655118\_\allowbreak{}006.  Build an 8 $\times$ 8 $\times$ 8 unit cell system FCC copper crystal with periodic boundary conditions in all directions. First relax the perfect crystal using conjugate-gradient energy minimization and record the total potential energy and number of atoms. Then remove exactly one atom from the center of the crystal, relax the defective structure again using conjugate-gradient minimization, and record the final potential energy. Calculate the vacancy formation energy using the perfect- and defective-crystal energies, and print the final value in eV.\\[2pt]
\footnotesize 

\item \textit{Thermal Expansion}\\
Calculate the linear thermal expansion coefficient of BCC iron at 293 K. Use the KIM potential EAM\_\allowbreak{}Dynamo\_\allowbreak{}AcklandBaconCalder\_\allowbreak{}1997\_\allowbreak{}Fe\_\allowbreak{}\_\allowbreak{}MO\_\allowbreak{}142799717516\_\allowbreak{}006. Build a 10 x 10 x 10 unit cell system with periodic boundary conditions in all directions. Initialise velocities at 253 K, then equilibrate in the NPT ensemble at 253 K and zero pressure for 50 ps. Then run a further 50 ps at 253 K and time-average the x box length over that 50 ps only. Ramp to 293 K over 10 ps, equilibrate for 40 ps, then run a further 50 ps and average again; then ramp to 333 K over 10 ps, equilibrate for 40 ps, run a further 50 ps and average again. Calculate the linear thermal expansion coefficient at 293 K as (L333 - L253) / (L293 * 80.0), print it in K\textasciicircum{}-1 together with the three averaged box lengths.

\item \textit{Tensile Loading}\\
Simulate uniaxial tensile deformation of an FCC aluminum single crystal at 300 K. Use the KIM potential EAM\_\allowbreak{}Dynamo\_\allowbreak{}MishinFarkasMehl\_\allowbreak{}1999\_\allowbreak{}Al\_\allowbreak{}\_\allowbreak{}MO\_\allowbreak{}651801486679\_\allowbreak{}006. Build a 10 $\times$ 10 $\times$ 10 conventional-unit-cell crystal with periodic boundary conditions. Equilibrate the system at 300 K and 1 bar in the NPT ensemble for 20 ps, then apply tensile deformation along the x direction at a constant engineering strain rate of 10\textasciicircum{}10/s for 20 ps. Maintain 1 bar lateral pressure in the y and z directions during loading. Output the engineering strain together with the normal stress components every 100 steps.

\item \textit{Cyclic Loading}\\
Simulate crack propagation in FCC copper at 300 K. Use the KIM potential EAM\_\allowbreak{}Dynamo\_\allowbreak{}MishinMehlPapaconstantopoulos\_\allowbreak{}2001\_\allowbreak{}Cu\_\allowbreak{}\_\allowbreak{}MO\_\allowbreak{}346334655118\_\allowbreak{}006. First equilibrate a 20x20x5 unit cell system using NPT for 5 ps at 1 bar pressure in the x, y, and z directions. Then create a centered horizontal crack by deleting atoms: the crack lies in the xz plane at the centre of the cell, with a length along x equal to 20\% of the box length in x and an opening along y of two lattice parameters.  NVT relax at 300 K for 5 ps. Then apply cyclic tensile loading along y under NVT at 300 K, with a strain amplitude of 2\% and a period of 10 ps, for a total of 20 ps. Print thermodynamic output every 100 steps including the yy component of the pressure tensor.
\end{enumerate}

\section{LAMMPS scripting rules}
\label{sup:rules}

The rules below are the block added to the knowledge base in the third condition of the script-generation
benchmark in the main text. They were written by reading the errors LAMMPS returned on
ChatAtom's own generated scripts, and are supplied verbatim to both the generator and the
reviewer of the LAMMPS agent. They are reproduced here as the system receives them.

\begin{lstlisting}[basicstyle=\ttfamily\scriptsize,breaklines=true,frame=single,
                   columns=fullflexible,keepspaces=true,showstringspaces=false,numbers=none]
LAMMPS SCRIPTING RULES:

ORDERING
- Define every variable, compute and fix ABOVE its first reference. LAMMPS parses
  top to bottom.

- 'variable X equal <expr>' stores the FORMULA, not a number, and re-evaluates on
  every read -- including 'f_ID' and 'c_ID', which also fail once the fix or compute
  is gone. Freeze it where the measurement is taken: 'variable X equal $(<expr>)'.
  $() evaluates immediately, so a compute needs a run above it.

UNITS
- region, create_atoms and fix deform default to LATTICE units. Anything derived
  from lx/ly/lz or a lattice constant needs 'units box', or it silently matches
  the wrong atoms.

CRYSTAL AND BAROSTAT
- fcc: Al Cu Ni Ag Au | bcc: Fe W Mo Nb Cr | hcp: Ti Zn Mg Zr Co | diamond: Si Ge C
- A non-cubic cell needs 'aniso', 'tri', or independent x/y/z on the barostat.
  'iso' scales all axes together and FREEZES the cell shape

BOUNDARIES
- Non-periodic directions need 's', or 'f' WITH vacuum padding. 'f' with the box
  flush against the crystal aborts with "ERROR: Lost atoms" on the first step

MODIFYING THE SYSTEM
- Use 'delete_atoms ... compress no' to keep atom IDs stable across phases.
- No need to call 'velocity create' afterwards: 'loop all' requires contiguous IDs, and
  the surviving atoms already carry equilibrated velocities.

DEFORMATION AND BAROSTAT
- 'fix deform' and the barostat must not control the SAME box dimension. LAMMPS
  aborts with "Must not have multiple fixes change box parameter x".
- Barostatting the OTHER axes is correct and required for uniaxial STRESS, i.e. a
  real tensile test where the lateral axes contract by Poisson:
      fix load all npt temp ${temp} ${temp} 0.1 y 1.0 1.0 1.0 z 1.0 1.0 1.0
      fix pull all deform 1 x erate 0.01 units box

SUBSTITUTION
- '${name}' reads a VARIABLE you defined; '$(expr)' evaluates an expression NOW.
  Thermo keywords are NOT variables: '${lx}', '${step}', '${press}' abort the run.
  Use '$(lx)'. '${v_name}' is wrong too -- it
  looks for a variable literally called "v_name".

COMPUTES
- A compute has no value until a run starts it.
  'dump ... c_ID' then 'run N' is fine: the run starts the compute, and the
  dump saves step 0, which is the starting structure.
  'write_dump ... c_ID' and 'variable X equal c_ID' fail if no run has happened
  yet; after a run they work. To save the starting structure use the dump's
  step 0, not write_dump. To read a compute before any run, put 'run 0' above it.

- 'lattice hcp <a>' always builds the IDEAL c/a = 1.63299 and has NO c/a keyword
\end{lstlisting}

\section{Physical correctness of the benchmark scripts}
\label{sup:correctness}

Tables~\ref{tab:slattice}, \ref{tab:svacancy} and~\ref{tab:sthermal} list the value returned by every generated script for the three benchmark prompts whose requested quantity OpenKIM reports
for the same interatomic potential. Each prompt was issued three times to each of six
models under each of the three conditions, giving 18 scripts per condition.

\begin{table}[H]
\centering
\footnotesize
\caption{Equilibrium lattice parameter of fcc aluminum, in \AA{}, returned by every generated script for the lattice-parameter prompt. Entries marked -- did not execute. The OpenKIM reference value for this potential is 4.05~\AA{}.}
\label{tab:slattice}
\begin{tabular}{@{}llrrr@{}}
\toprule
Model & Rep. & Zero-shot & ChatAtom & ChatAtom + rules \\
\midrule
Opus 4.6 & r0 & -- & 4.05 & 4.05 \\
 & r1 & -- & -- & 4.05 \\
 & r2 & -- & 4.05 & 4.05 \\
Opus 4.8 & r0 & 4.05 & 4.05 & 4.05 \\
 & r1 & 4.05 & 4.05 & 4.05 \\
 & r2 & 4.05 & 4.05 & 4.05 \\
Opus 5 & r0 & 4.05 & 4.05 & 4.05 \\
 & r1 & 4.05 & 4.05 & 4.05 \\
 & r2 & 4.05 & 4.05 & 4.05 \\
GPT-5.4 & r0 & -- & 4.05 & 4.05 \\
 & r1 & -- & 4.05 & 4.05 \\
 & r2 & -- & -- & 4.05 \\
GPT-5.5 & r0 & -- & 4.05 & 4.05 \\
 & r1 & -- & 4.05 & 4.05 \\
 & r2 & 4.05 & 4.05 & 4.05 \\
GPT-5.6 & r0 & 4.05 & 4.05 & 4.05 \\
 & r1 & 4.05 & 4.05 & 4.05 \\
 & r2 & 4.05 & 4.05 & 4.05 \\
\midrule
\multicolumn{2}{@{}l}{Executed} & 10 / 18 & 16 / 18 & 18 / 18 \\
\multicolumn{2}{@{}l}{Correct}  & 10 & 16 & 18 \\
\bottomrule
\end{tabular}
\end{table}

\begin{table}[H]
\centering
\footnotesize
\caption{Monovacancy formation energy of fcc copper, in eV, returned by every generated script for the vacancy prompt. Entries marked -- did not execute. The OpenKIM reference value for this potential is 1.27225~eV.}
\label{tab:svacancy}
\begin{tabular}{@{}llrrr@{}}
\toprule
Model & Rep. & Zero-shot & ChatAtom & ChatAtom + rules \\
\midrule
Opus 4.6 & r0 & 1.2722 & 1.2724 & 1.2722 \\
 & r1 & -- & 1.2724 & 1.2724 \\
 & r2 & 1.2724 & 1.2724 & 1.2722 \\
Opus 4.8 & r0 & $-3.5394$ & 1.2725 & 1.2707 \\
 & r1 & 1.2724 & -- & 1.2724 \\
 & r2 & $-3.5394$ & 1.2725 & -- \\
Opus 5 & r0 & 1.2724 & 1.2724 & 1.2724 \\
 & r1 & $-3.5400$ & 1.2724 & 1.2724 \\
 & r2 & 0 & 1.2724 & 1.2724 \\
GPT-5.4 & r0 & 0 & $-3.5394$ & 1.2724 \\
 & r1 & 0 & -- & 1.2724 \\
 & r2 & 0 & 1.2724 & 1.2723 \\
GPT-5.5 & r0 & 1.2724 & 1.2724 & 1.2724 \\
 & r1 & 1.2724 & 1.2724 & 1.2724 \\
 & r2 & 1.2724 & 1.2724 & 1.2724 \\
GPT-5.6 & r0 & -- & -- & -- \\
 & r1 & -- & 1.2724 & 1.2724 \\
 & r2 & 1.2724 & 1.2724 & 1.2724 \\
\midrule
\multicolumn{2}{@{}l}{Executed} & 15 / 18 & 15 / 18 & 16 / 18 \\
\multicolumn{2}{@{}l}{Correct}  & 8 & 14 & 16 \\
\bottomrule
\end{tabular}
\end{table}

\begin{table}[H]
\centering
\footnotesize
\caption{Linear thermal expansion coefficient of bcc iron at 293~K, in K$^{-1}$, returned by every generated script for the thermal-expansion prompt. Entries marked -- did not execute. OpenKIM reports 9.5673$\times$10$^{-6}$~K$^{-1}$ for this potential at 293.15~K. A value counts as correct if it lies within 5\% of that reference, a wider band than for the other two prompts because the coefficient is obtained from a finite difference of time-averaged box lengths.}
\label{tab:sthermal}
\begin{tabular}{@{}llrrr@{}}
\toprule
Model & Rep. & Zero-shot & ChatAtom & ChatAtom + rules \\
\midrule
Opus 4.6  & r0 & 0 & 9.4978e-06 & 9.4978e-06 \\
          & r1 & -- & 9.3391e-06 & 9.4570e-06 \\
          & r2 & -- & -- & 9.4250e-06 \\
Opus 4.8  & r0 & -- & -- & 9.3829e-06 \\
          & r1 & -- & 9.3913e-06 & 9.3995e-06 \\
          & r2 & -- & -- & 9.4017e-06 \\
Opus 5    & r0 & 9.3883e-06 & 9.4161e-06 & 9.4650e-06 \\
          & r1 & -- & 9.4085e-06 & 9.4584e-06 \\
          & r2 & 9.3883e-06 & 9.4133e-06 & 9.3575e-06 \\
GPT-5.4   & r0 & -- & 9.4017e-06 & 9.8575e-06 \\
          & r1 & -- & 9.4976e-06 & 9.4230e-06 \\
          & r2 & -- & 9.3996e-06 & 9.3996e-06 \\
GPT-5.5   & r0 & 9.3865e-06 & 9.4976e-06 & 9.3680e-06 \\
          & r1 & 9.4107e-06 & 9.2529e-06 & 9.4241e-06 \\
          & r2 & -- & 9.4076e-06 & 8.2869e-06 \\
GPT-5.6   & r0 & 9.3989e-06 & -- & 9.4241e-06 \\
          & r1 & 9.3874e-06 & 9.0422e-06 & 9.4131e-06 \\
          & r2 & 9.4890e-06 & 9.4348e-06 & -- \\
\midrule
\multicolumn{2}{@{}l}{Executed} & 8 / 18 & 14 / 18 & 17 / 18 \\
\multicolumn{2}{@{}l}{Within 5\%} & 7 & 13 & 16 \\
\bottomrule
\end{tabular}
\end{table}

\section{Generated scripts}
\label{sup:scripts}

The two LAMMPS input scripts below were produced by the LAMMPS agent from the natural-language
requests described in the main text, after the user approved the proposed plan. They are
reproduced verbatim, with no manual editing. The copper and aluminum scripts differ only in
the interatomic model, the lattice parameter, the atomic mass, the common-neighbor-analysis
cutoff, and the velocity seed; the phase structure, crack geometry, and loading protocol are
identical.

\lstset{
  basicstyle=\ttfamily\scriptsize,
  breaklines=true,
  frame=single,
  columns=fullflexible,
  keepspaces=true,
  showstringspaces=false,
  captionpos=b,
  numbers=left,
  numberstyle=\tiny\color{gray},
  commentstyle=\color{blue!55!black},
}

\begin{lstlisting}[caption={Agent-generated LAMMPS input script for the Griffith-crack
simulation in copper.},label={lst:cu}]
# =============================================================
# Griffith Crack with Monotonic Loading - Cu
# KIM Model: EAM_Dynamo_MishinMehlPapaconstantopoulos_2001_Cu__MO_346334655118_006
# Total: 120 ps (10 ps pre-stress + 10 ps crack relaxation + 100 ps loading)
# =============================================================

# --- Initialize ---
units metal
dimension 3
boundary p p p
atom_style atomic

# --- KIM Init (BEFORE create_box) ---
kim init EAM_Dynamo_MishinMehlPapaconstantopoulos_2001_Cu__MO_346334655118_006 metal

# --- Variables ---
variable a0 equal 3.615
variable temp equal 300.0
variable timestep_val equal 0.001
variable thermo_freq equal 100
variable dump_freq equal 4000

# --- Create System: 50 x 50 x 10 unit cells ---
lattice fcc ${a0}
region simbox block 0 50 0 50 0 10 units lattice
create_box 1 simbox
create_atoms 1 box

# --- KIM Interactions (AFTER create_box) ---
kim interactions Cu

# --- Set Mass (REQUIRED) ---
mass 1 63.546

# --- Timestep ---
timestep ${timestep_val}

# --- Computes for crack analysis ---
compute peratom all pe/atom
compute stress_atom all stress/atom NULL
compute cna all cna/atom 3.086

# --- Initial Velocities ---
velocity all create ${temp} 87654 dist gaussian mom yes rot yes

# =============================================================
# PHASE 1: NPT Pre-stress Equilibration (10 ps, 10000 steps)
# =============================================================
print "========== PHASE 1: NPT Pre-stress Equilibration =========="

fix npt_equil all npt temp ${temp} ${temp} 0.1 x 1.0 1.0 1.0 y 1.0 1.0 1.0 z 1.0 1.0 1.0

thermo ${thermo_freq}
thermo_style custom step time temp pe press pxx pyy pzz vol lx ly lz

dump prestressDump all custom ${dump_freq} dump.*.prestress id type x y z c_peratom c_stress_atom[1] c_stress_atom[2] c_stress_atom[3] c_cna

run 10000

undump prestressDump
unfix npt_equil

# =============================================================
# PHASE 2: Crack Seeding (delete_atoms) + NVT Relaxation (10 ps, 10000 steps)
# =============================================================
print "========== PHASE 2: Crack Seeding + NVT Relaxation =========="

# Define crack geometry after NPT equilibration (relaxed box dimensions)
# Crack: 10% of Lx in length, centered at box center, 1 a0 wide in y
variable xlo_crack equal lx/2-0.05*lx
variable xhi_crack equal lx/2+0.05*lx
variable ylo_crack equal ly/2-0.5*${a0}
variable yhi_crack equal ly/2+0.5*${a0}

# Create crack region and delete atoms
region crack_slot block ${xlo_crack} ${xhi_crack} ${ylo_crack} ${yhi_crack} INF INF units box
delete_atoms region crack_slot compress no

# NVT relaxation around crack
fix nvt_relax all nvt temp ${temp} ${temp} 0.1

dump relaxDump all custom ${dump_freq} dump.*.relaxation id type x y z c_peratom c_stress_atom[1] c_stress_atom[2] c_stress_atom[3] c_cna

run 10000

undump relaxDump
unfix nvt_relax

# =============================================================
# PHASE 3: Monotonic Tensile Loading in y (100 ps, 100000 steps)
# Strain rate: 1e-3 per ps in y-direction (Mode I opening)
# =============================================================
print "========== PHASE 3: Monotonic Tensile Loading =========="

fix nvt_load all nvt temp ${temp} ${temp} 0.1
fix deform_load all deform 1 y erate 0.001 units box remap x

dump loadDump all custom ${dump_freq} dump.*.loading id type x y z c_peratom c_stress_atom[1] c_stress_atom[2] c_stress_atom[3] c_cna

run 100000

undump loadDump
unfix deform_load
unfix nvt_load

# =============================================================
print "========== Simulation Complete =========="
print "Total steps: 120000 (10000 + 10000 + 100000)"
print "Total time: 120 ps"
\end{lstlisting}

\begin{lstlisting}[caption={Agent-generated LAMMPS input script for the Griffith-crack
simulation in aluminum.},label={lst:al}]
# =============================================================
# Griffith Crack Simulation in Al (Monotonic Loading)
# KIM Model: EAM_Dynamo_MishinFarkasMehl_1999_Al__MO_651801486679_006
# Temperature: 300 K
# System: 50 x 50 x 10 unit cells
# Total: 120 ps (120,000 steps)
# Phase 1: NPT pre-stress equilibration (10 ps, 10,000 steps)
# Phase 2: Crack seeding + NVT relaxation (10 ps, 10,000 steps)
# Phase 3: Monotonic tensile loading in Y (100 ps, 100,000 steps)
# =============================================================

# --- Initialization ---
kim init EAM_Dynamo_MishinFarkasMehl_1999_Al__MO_651801486679_006 metal

# --- Variables ---
variable a0 equal 4.05
variable temp equal 300.0
variable timestep_val equal 0.001
variable thermo_freq equal 100
variable dump_freq equal 4000
variable cna_cutoff equal 3.457

# --- Create simulation box ---
boundary p p p
lattice fcc ${a0}
region simbox block 0 50 0 50 0 10 units lattice
create_box 1 simbox
create_atoms 1 box

# --- Force field ---
kim interactions Al
mass 1 26.9815

# --- Computes for output ---
compute peratom all pe/atom
compute stress_atom all stress/atom NULL
compute cna all cna/atom ${cna_cutoff}

# --- Simulation settings ---
timestep ${timestep_val}
thermo ${thermo_freq}
thermo_style custom step time temp pe press pxx pyy pzz vol lx ly lz

# --- Initial velocities ---
velocity all create ${temp} 12345 mom yes rot yes

# =============================================================
# PHASE 1: NPT Pre-stress Equilibration (10 ps, 10,000 steps)
# Px = 1.0 bar, Py = 1.0 bar, Pz = 1.0 bar
# =============================================================
fix npt_equil all npt temp ${temp} ${temp} 0.1 x 1.0 1.0 1.0 y 1.0 1.0 1.0 z 1.0 1.0 1.0

dump phase1dump all custom ${dump_freq} dump.phase1.* id type x y z c_peratom c_stress_atom[1] c_stress_atom[2] c_stress_atom[3] c_cna

run 10000

undump phase1dump
unfix npt_equil

# =============================================================
# PHASE 2: Crack Seeding + NVT Relaxation (10 ps, 10,000 steps)
# Crack: centered at Y=Ly/2, length=10% of Lx, width=1 a0 in Y
# Method: delete_atoms
# =============================================================

# Define crack geometry using current (relaxed) box dimensions
variable xlo_crack equal lx/2-0.05*lx
variable xhi_crack equal lx/2+0.05*lx
variable ylo_crack equal ly/2-0.5*${a0}
variable yhi_crack equal ly/2+0.5*${a0}

# Create crack region and delete atoms
region crack_slot block ${xlo_crack} ${xhi_crack} ${ylo_crack} ${yhi_crack} INF INF units box
delete_atoms region crack_slot compress no

# NVT relaxation of cracked system
fix nvt_relax all nvt temp ${temp} ${temp} 0.1

dump phase2dump all custom ${dump_freq} dump.phase2.* id type x y z c_peratom c_stress_atom[1] c_stress_atom[2] c_stress_atom[3] c_cna

run 10000

undump phase2dump
unfix nvt_relax

# =============================================================
# PHASE 3: Monotonic Tensile Loading in Y (100 ps, 100,000 steps)
# Strain rate: 1e-3 per ps (erate 0.001)
# Mode I crack opening
# =============================================================
fix nvt_load all nvt temp ${temp} ${temp} 0.1
fix deform_load all deform 1 y erate 0.001 units box

dump phase3dump all custom ${dump_freq} dump.phase3.* id type x y z c_peratom c_stress_atom[1] c_stress_atom[2] c_stress_atom[3] c_cna

run 100000

undump phase3dump
unfix deform_load
unfix nvt_load

# --- Done ---
print "Simulation complete: 120 ps total (120,000 steps)"
\end{lstlisting}

\section{Analysis agent output}
\label{sup:agent-output}

The four reports below are the output of the analysis agent for the two Griffith-crack runs
described in the main text. They are reproduced as returned, with formatting applied to the
tables and no change to the wording. Each was produced from the run directory alone, after
the simulation had completed.

\subsection{Copper --- stress--strain analysis}

\noindent\textit{Analysis: Cu Griffith Crack --- Monotonic Tensile Loading}

\textbf{System:} 100,000-atom FCC copper slab ($50 \times 50 \times 10$ unit cells,
$a_0 = 3.615$~\AA), Mishin EAM potential (2001), 300~K, with a pre-seeded central crack
($\approx 10\%$ of $L_x$, 200 atoms removed).

\textbf{Simulation protocol (120 ps total):}
\begin{enumerate}[leftmargin=2em,itemsep=2pt]
\item \textbf{Phase 1} --- 10 ps NPT pre-stress equilibration at 300~K and 1 bar.
\item \textbf{Phase 2} --- 10 ps NVT relaxation after crack insertion (constant volume).
\item \textbf{Phase 3} --- 100 ps uniaxial tensile loading along y at
$\dot{\varepsilon} = 10^{9}$~s$^{-1}$ (Mode I opening).
\end{enumerate}

\textbf{Key Results}

\noindent\begin{tabular}{@{}lp{10.5cm}@{}}
\toprule
Quantity & Value \\
\midrule
Peak (yield) stress & 8.68 GPa at $\varepsilon \approx 0.063$ \\
Pre-yield response  & Nearly linear rise from 0 to $\approx 8.6$ GPa over 0--6.3\% strain \\
Post-peak drop      & Sharp fall: 8.65 to 5.88 GPa within $\Delta\varepsilon \approx 0.01$
                      ($\approx 32\%$ stress drop) \\
Residual plateau    & Stress settles to $\approx 3.1$--3.5 GPa from $\varepsilon \approx 0.08$
                      onward, remaining there through 10\% strain \\
Temperature         & Well-controlled at $\approx 300$~K throughout all phases \\
Equilibration       & Phase 1 converged ($T$, $P$, PE all stable by $\approx 3$ ps); Phase 2
                      NVT relaxation stable with a slight negative mean pressure
                      ($\approx -280$ bar) due to the free surfaces of the crack \\
\bottomrule
\end{tabular}

\medskip
\textbf{Mechanical Interpretation}

The stress-strain curve shows classic crack-tip yielding behavior in a ductile FCC metal:

\begin{itemize}[leftmargin=2em,itemsep=2pt]
\item The linear elastic regime extends to $\approx 6\%$ strain, consistent with the high
strain rate and nanoscale crack geometry concentrating stress at the crack tip.
\item The sharp stress drop at $\varepsilon \approx 0.063$ signals the onset of crack-tip
plasticity --- likely dislocation emission from the crack tip relieving the stress
concentration. The drop is abrupt ($\approx 3$ GPa over $\approx 1\%$ strain), indicating a
sudden burst of plastic activity rather than gradual yielding.
\item The post-yield plateau at $\approx 3.2$ GPa indicates the crack has blunted and the
material continues to deform plastically at a roughly constant flow stress, characteristic of
ductile failure by crack-tip blunting rather than brittle cleavage. The stress does not
collapse to zero, confirming the crack did not propagate catastrophically.
\end{itemize}

\textbf{In summary:} This copper slab with a Griffith crack exhibits a ductile fracture
response --- elastic loading to $\approx 8.7$ GPa followed by crack-tip plasticity and
blunting, with a sustained flow stress of $\approx 3.2$ GPa. If you'd like to examine the
dislocation mechanism (partial vs.\ full dislocations, nucleation timing), just ask and I can
run the DXA analysis.

\subsection{Copper --- dislocation analysis}

\textbf{Nucleation Onset.} The crystal remained defect-free for 72\% of the simulation
(steps 0--84{,}000). Dislocations first nucleated at step 88{,}000 with an initial density of
$\approx 8.1 \times 10^{15}$~m$^{-2}$, consisting of just 6 dislocation lines --- all Shockley
$\frac{1}{6}\langle 112 \rangle$ partials.

\textbf{Multiplication and Peak.} Density climbed $\times 36$ from onset to a peak of
$\approx 2.95 \times 10^{17}$~m$^{-2}$ at step 116{,}000 (218 lines, $\approx 3870$~\AA{} total
line length). The multiplication rate was $\approx 1.0 \times 10^{13}$~m$^{-2}$ per timestep ---
rapid, consistent with a burst of dislocation emission from a crack tip or stress concentrator.

\medskip
\noindent\textbf{Burgers-Vector Family Breakdown (at peak)}

\noindent\begin{tabular}{@{}lll@{}}
\toprule
Family & Fraction & Line length \\
\midrule
Shockley $\frac{1}{6}\langle 112 \rangle$  & 76\% & 2953~\AA \\
Stair-rod $\frac{1}{6}\langle 110 \rangle$ & 11\% & 415~\AA \\
Hirth $\frac{1}{3}\langle 100 \rangle$     & 3\%  & 130~\AA \\
Other                                       & 8\%  & 322~\AA \\
\bottomrule
\end{tabular}

\medskip
\textbf{Post-Peak Behaviour.} Density dropped only $\approx 3\%$ from peak to the final frame
(step 120{,}000), indicating mild annihilation or absorption at free surfaces rather than a
dramatic recovery event.

\textbf{Structural Composition (final frame).} 82\% FCC, 9.5\% HCP (stacking-fault planes),
8.3\% Other (crack faces, surfaces, cores).

\textbf{Interpretation.} The deformation is partial-dislocation-mediated plasticity. The
overwhelming dominance of Shockley partials (73--76\%) together with the substantial HCP
fraction ($\approx 9.5\%$) confirms that leading partials glide away from the crack tip, leaving
behind intrinsic stacking faults on \{111\} planes --- the classic ductile blunting mechanism in
FCC copper.

The 11\% stair-rod $\frac{1}{6}\langle 110 \rangle$ content signals the formation of
Lomer-Cottrell locks where partials on intersecting \{111\} planes react. These sessile locks
impede further glide and contribute to strain hardening, which is why the density plateaus
rather than continuing to climb.

If you'd like, I can also pull the stress--strain curve for this run so we can cross-check
whether the yield point coincides with the dislocation nucleation onset at step 88{,}000.

\subsection{Aluminum --- stress--strain analysis}

\noindent\textit{Analysis: Griffith Crack Simulation in Aluminum}

\textbf{System Identification.} Material: Aluminum (FCC), Mishin EAM potential (1999).
System size: $50 \times 50 \times 10$ unit cells (100,000 atoms; 200 deleted for crack
$\rightarrow$ 99,800). Temperature: 300~K (NVT during loading). Crack: central slot, 10\% of
$L_x$ long, 1 $a_0$ wide, oriented for Mode I opening in Y. Loading: monotonic tensile strain
in Y at $\dot{\varepsilon} = 10^{9}$~s$^{-1}$ for 100 ps (10\% total strain).

\medskip
\noindent\textbf{Three-Phase Protocol}

\noindent\begin{tabular}{@{}llll@{}}
\toprule
Phase & Duration & Ensemble & Purpose \\
\midrule
1 & 0--10 ps   & NPT                       & Stress-free equilibration at 300~K, 1 bar \\
2 & 10--20 ps  & NVT                       & Crack seeding + relaxation (fixed box) \\
3 & 20--120 ps & NVT + \texttt{fix deform} & Uniaxial tensile loading in Y \\
\bottomrule
\end{tabular}

\medskip
Phase 1 equilibrated well: temperature settled to $\approx 300$~K and pressure fluctuated
around zero within $\pm 400$ bar --- normal for a 100k-atom NPT run. Phase 2 relaxed the
cracked system at constant volume. Temperature remained stable at $\approx 300$~K and the box
dimensions were frozen, confirming NVT behavior. The potential energy dropped by
$\approx 770$ eV upon crack creation (from $-332{,}135$ to $-331{,}366$ eV), consistent with the
loss of 200 atoms and the creation of free surfaces.

\textbf{Mechanical Response (Stress--Strain).} The stress--strain curve reveals a clear
elastic--plastic--failure sequence:

\begin{itemize}[leftmargin=2em,itemsep=2pt]
\item \textbf{Linear elastic regime} ($\varepsilon \approx 0$--0.04): stress rises nearly
linearly from 0 to $\approx 4$ GPa. The response is stiff and monotonic.
\item \textbf{Peak stress:} $\sigma_{\mathrm{peak}} \approx 5.04$ GPa at
$\varepsilon \approx 0.057$. This marks the onset of crack propagation / plastic yielding from
the crack tip.
\item \textbf{Post-peak softening} ($\varepsilon \approx 0.057$--0.075): stress drops sharply
from 5.0 GPa to $\approx 2.3$ GPa --- a rapid $\approx 55\%$ reduction over just $\approx 2\%$
strain. This is characteristic of catastrophic crack propagation rather than gradual ductile
necking.
\item \textbf{Residual plateau and gradual decline} ($\varepsilon \approx 0.075$--0.10): stress
continues to decrease from $\approx 2.3$ GPa to $\approx 1.1$ GPa, declining steadily rather
than collapsing to zero. This indicates the crack has propagated significantly but the sample
has not fully separated --- residual load-bearing ligaments persist, and the material continues
to deform plastically at the crack tip.
\end{itemize}

\textbf{Energy and Pressure Trends During Loading.} Potential energy decreased steadily from
$\approx -331{,}370$ eV (step 20{,}000) to $\approx -329{,}580$ eV (step 120{,}000), a rise of
$\approx 1790$ eV, reflecting elastic strain energy storage followed by energy release during
crack propagation. $P_{yy}$ grew from near zero to $\approx -50{,}400$ bar
($\approx -5.04$ GPa) at peak, then relaxed back toward $\approx -10{,}800$ bar by the end ---
consistent with the stress--strain curve. Temperature remained well-controlled at
$300 \pm 2$~K throughout, confirming the NVT thermostat functioned properly even during
fracture.

\textbf{Scientific Interpretation.} This simulation captures Mode I crack opening in FCC
aluminum at 300~K. The peak stress of $\approx 5.0$ GPa is high compared to bulk experimental
yield ($\approx 0.1$--0.3 GPa) because the system is nanoscale, defect-free except for the seed
crack, and loaded at an extreme strain rate ($10^{9}$~s$^{-1}$). The sharp post-peak stress drop
indicates rapid crack propagation once the critical stress intensity is reached, while the
non-zero residual stress ($\approx 1$ GPa at 10\% strain) suggests the crack has not fully
traversed the sample and plastic deformation at the crack tip is absorbing energy --- consistent
with aluminum's intrinsically ductile character even under dynamic fracture conditions.

\subsection{Aluminum --- dislocation analysis}

\textbf{System:} Aluminium (FCC), Mishin EAM potential (1999), 100,000 atoms
($50 \times 50 \times 10$ unit cells), 300~K. Three-phase simulation: NPT equilibration
$\rightarrow$ crack seeding + NVT relaxation $\rightarrow$ monotonic tensile loading in Y at
$\dot{\varepsilon} = 10^{9}$~s$^{-1}$.

\medskip
\noindent\textbf{Key DXA Results}

\noindent\begin{tabular}{@{}lp{10.5cm}@{}}
\toprule
Quantity & Value \\
\midrule
Nucleation onset       & Step 76{,}000 ($t = 76$ ps, $\varepsilon_y \approx 5.6\%$) --- first
                         dislocation line detected \\
Defect-free fraction   & 62\% of frames before onset --- the crystal remained pristine through
                         most of the loading \\
First dislocation type & 100\% Shockley $\frac{1}{6}\langle 112 \rangle$ partial
                         (single 17~\AA{} line) \\
Peak density           & $9.9 \times 10^{16}$~m$^{-2}$ at step 116{,}000 ($t = 116$ ps,
                         $\varepsilon_y \approx 9.6\%$) --- 103 dislocation lines, 1819~\AA{}
                         total line length \\
Final density          & $8.0 \times 10^{16}$~m$^{-2}$ at step 120{,}000 --- 19\% drop from
                         peak \\
Multiplication rate    & $\times 106$ from onset to peak over 40{,}000 steps
                         ($\approx 2.5 \times 10^{12}$~m$^{-2}$ per step) \\
\bottomrule
\end{tabular}

\medskip
\noindent\textbf{Burgers-Vector Family Breakdown}

\noindent\begin{tabular}{@{}lll@{}}
\toprule
Family & At Peak (step 116k) & At Final (step 120k) \\
\midrule
Shockley $\frac{1}{6}\langle 112 \rangle$  & 66\% (1196~\AA) & 73\% (1079~\AA) \\
Stair-rod $\frac{1}{6}\langle 110 \rangle$ & 22\% (395~\AA)  & 15\% (214~\AA) \\
Perfect $\frac{1}{2}\langle 110 \rangle$   & 7\% (131~\AA)   & 4\% (56~\AA) \\
Other                                       & 4\%             & 6\% \\
\bottomrule
\end{tabular}

\medskip
The final structure is 92.5\% FCC, 0.7\% HCP (stacking-fault planes), and 6.8\% ``Other''
(crack faces, free surfaces, dislocation cores).

\textbf{Interpretation.}

\begin{itemize}[leftmargin=2em,itemsep=2pt]
\item \textbf{Partial-mediated plasticity:} the overwhelming dominance of Shockley
$\frac{1}{6}\langle 112 \rangle$ partials --- both as the first nucleated defect and as the
majority family throughout --- confirms that crack-tip plasticity in this Al system proceeds via
emission of leading partial dislocations that trail stacking faults (consistent with the 0.7\%
HCP fraction, which marks the fault ribbons).
\item \textbf{Lomer-Cottrell lock formation:} the significant stair-rod
$\frac{1}{6}\langle 110 \rangle$ content (peaking at 22\%) indicates that Shockley partials on
intersecting \{111\} planes reacted to form sessile Lomer-Cottrell locks. These locks impede
further dislocation glide and contribute to strain hardening --- explaining why the stress
continued to build even after initial dislocation nucleation.
\item \textbf{Late nucleation onset:} dislocations did not appear until step 76{,}000
($\approx 5.6\%$ strain), meaning the crack tip sustained purely elastic stress concentration for
a prolonged period before emitting the first partial. This is characteristic of a ductile
blunting response: the crack did not propagate by brittle cleavage but instead relieved the
stress concentration through dislocation emission.
\item \textbf{Post-peak annihilation:} the 19\% density drop from peak to final frame over the
last 4000 steps suggests dislocations are being absorbed at the free surfaces (the thin
z-dimension of 10 unit cells provides nearby sinks), or partial annihilation events are
occurring as the stress relaxes at high strain.
\end{itemize}

In summary, this Al crack simulation exhibits a ductile, partial-dislocation-mediated fracture
mechanism with significant stacking-fault activity and Lomer-Cottrell hardening, consistent with
the moderate stacking-fault energy of aluminium under the Mishin EAM potential.

\section{Autonomous campaign: template and visualization scripts}
\label{sup:campaign-template}

The campaign runs from the single LAMMPS input script below. The LAMMPS agent wrote
it on the first trial, through the same generate--review loop it uses for any other
simulation, and the user approved it once before the search began. The three swept parameters are
declared in the block at the top; at each iteration the campaign substitutes the
proposed values into that block and leaves the rest of the script untouched, so that
a difference between trials can be attributed to the parameters under test and not
to a rewritten script. The crack is seeded by excluding the neighbor interactions
between the two groups of atoms facing each other across the crack plane, leaving
the atoms in place.

\begin{lstlisting}[basicstyle=\ttfamily\scriptsize,breaklines=true,frame=single,
columns=fullflexible,keepspaces=true,showstringspaces=false,captionpos=b,
numbers=left,numberstyle=\tiny\color{gray},commentstyle=\color{blue!55!black},
caption={Campaign template script, as written by the LAMMPS agent and approved by the user.
The values shown in the swept block are those of a single iteration.},
label={lst:campaign}]
# LAMMPS input script: Cu crack + monotonic tensile loading campaign

# ==== CAMPAIGN VARIABLES (auto-filled per iteration — do not move) ====
# --- SWEPT KNOBS — campaign_fill_template rewrites these every iteration ---
variable temperature       equal 400
variable strain_rate       equal 0.01
variable crack_length_frac equal 0.3
# --- FIXED for this campaign — edit once at approval, never swept ---
variable a0                equal 3.615
variable crack_width_a0    equal 4
variable timestep_ps       equal 0.001
variable dump_freq         equal 1500
variable cna_cutoff        equal 3.087
# ==== END CAMPAIGN VARIABLES ====

# Derived variables
variable steps_phase1 equal 5000
variable steps_phase2 equal 5000
variable steps_phase3 equal 15000
variable tdamp equal 0.1
variable pdamp equal 1.0

# Initialize KIM model
kim init EAM_Dynamo_MishinMehlPapaconstantopoulos_2001_Cu__MO_346334655118_006 metal

# Setup simulation box
boundary p p p
lattice fcc ${a0}
region simbox block 0 60 0 30 0 3 units lattice
create_box 1 simbox
create_atoms 1 box

# Setup interactions
kim interactions Cu
mass 1 63.546

# Timestep
timestep ${timestep_ps}

# Velocity initialization
velocity all create ${temperature} 12345 mom yes rot yes dist gaussian

# Computes for output
compute peratom all pe/atom
compute stress_atom all stress/atom NULL
compute cna all cna/atom ${cna_cutoff}

# Thermo output
thermo ${dump_freq}
thermo_style custom step time temp pe press pxx pyy pzz vol lx ly lz

# ============================================================
# Phase 1: Pre-stress equilibration (NPT, 5 ps)
# ============================================================
fix npt_equil all npt temp ${temperature} ${temperature} ${tdamp} iso 1.0 1.0 ${pdamp}

dump equilDump all custom ${dump_freq} dump.*.equil id type x y z c_peratom c_stress_atom[1] c_stress_atom[2] c_stress_atom[3] c_cna

run ${steps_phase1}

undump equilDump
unfix npt_equil

# ============================================================
# Phase 2: Crack seeding + NVT relaxation (5 ps)
# ============================================================

# Define crack geometry after NPT equilibration (relaxed positions)
variable ymid equal ly/2
variable half_crack_len equal ${crack_length_frac}/2.0*lx
variable xbound_low equal lx/2-v_half_crack_len
variable xbound_high equal lx/2+v_half_crack_len
variable half_crack_wid equal ${crack_width_a0}/2.0*${a0}
variable ybound_low equal v_ymid-v_half_crack_wid
variable ybound_high equal v_ymid+v_half_crack_wid

# Create crack regions (horizontal crack at y=Ly/2, central along x)
region crack_lower block ${xbound_low} ${xbound_high} ${ybound_low} ${ymid} INF INF units box
region crack_upper block ${xbound_low} ${xbound_high} ${ymid} ${ybound_high} INF INF units box

# Create groups from regions
group crack_lower region crack_lower
group crack_upper region crack_upper

# Sever bonds across crack plane
neigh_modify exclude group crack_upper crack_lower

# NVT relaxation with crack
fix nvt_relax all nvt temp ${temperature} ${temperature} ${tdamp}

dump relaxDump all custom ${dump_freq} dump.*.relax id type x y z c_peratom c_stress_atom[1] c_stress_atom[2] c_stress_atom[3] c_cna

run ${steps_phase2}

undump relaxDump
unfix nvt_relax

# ============================================================
# Phase 3: Monotonic tensile loading (NVT, 15 ps)
# ============================================================
fix nvt_load all nvt temp ${temperature} ${temperature} ${tdamp}
fix deform_load all deform 1 y erate ${strain_rate} units box remap x

dump loadDump all custom ${dump_freq} dump.*.load id type x y z c_peratom c_stress_atom[1] c_stress_atom[2] c_stress_atom[3] c_cna

run ${steps_phase3}

undump loadDump
unfix deform_load
unfix nvt_load

print "Simulation complete: all three phases finished successfully"
\end{lstlisting}

The visualization below was written by the OVITO agent and is the same for every
simulation in the campaign. It reads the structure type that OVITO maps from the
per-atom common-neighbor analysis already computed in the LAMMPS run, folds the LAMMPS
encoding of unclassified atoms onto the OVITO one, and colors the atoms by structure
type. ChatAtom executes the script itself, so the pipeline object it appends to is
supplied by the executor rather than created in the script.

\begin{lstlisting}[language=Python,basicstyle=\ttfamily\scriptsize,breaklines=true,
frame=single,columns=fullflexible,keepspaces=true,showstringspaces=false,captionpos=b,
numbers=left,numberstyle=\tiny\color{gray},commentstyle=\color{blue!55!black},
caption={Agent-generated OVITO script used for every campaign trial.},
label={lst:campaign-ovito}]
from ovito.modifiers import PythonModifier
import numpy as np

def assign_cna_colors(frame, data):
    """Per-frame modifier: remap LAMMPS CNA values and assign custom colors."""
    
    cna_cutoff = 3.087
    
    # Read pre-computed Structure Type (auto-mapped from LAMMPS c_cna)
    cna_values = data.particles['Structure Type'][...]
    
    # Remap LAMMPS encoding: value 5 (Other) → 0 (OVITO Other)
    structure_type = np.where(cna_values >= 5, 0, cna_values).astype(int)
    
    # Write back remapped values using create_property
    data.particles_.create_property('Structure Type', data=structure_type)
    
    # Define colors per OVITO structure type:
    # 0 = Other → red, 1 = FCC → green, 2 = HCP → yellow, 3 = BCC → blue, 4 = ICO → gray
    color_map = {
        0: (1, 0, 0),      # Other = red
        1: (0, 1, 0),      # FCC = green
        2: (1, 1, 0),      # HCP = yellow
        3: (0, 0, 1),      # BCC = blue
        4: (0.5, 0.5, 0.5) # ICO = gray
    }
    
    num_particles = len(structure_type)
    colors = np.zeros((num_particles, 3), dtype=float)
    
    for type_id, rgb in color_map.items():
        mask = (structure_type == type_id)
        colors[mask] = rgb
    
    # Assign colors to particles
    data.particles_.create_property('Color', data=colors)

pipeline.modifiers.append(PythonModifier(function=assign_cna_colors))
\end{lstlisting}

\section{Autonomous campaign: proposed hypotheses}
\label{sup:campaign-hypotheses}

The campaign minimised the change in crack cavity surface area, $\Delta A$, over three
knobs: strain rate, temperature, and initial crack length as a fraction of the box length.
The table records, for each iteration, the knob values the orchestrator proposed, the
measured objective, and the reasoning it gave before the trial was run. The reasoning is
reproduced verbatim. Iteration~5 returned no measurable crack growth and was the best
trial of the campaign; the ten iterations that followed did not improve on it.

\begin{longtable}{@{}c c c c r p{0.50\textwidth}@{}}
\toprule
\# & $\dot{\varepsilon}$ (ps$^{-1}$) & $T$ (K) & $a/L_x$ & $\Delta A$ (\AA$^2$) & Proposed hypothesis \\
\midrule
\endfirsthead
\toprule
\# & $\dot{\varepsilon}$ (ps$^{-1}$) & $T$ (K) & $a/L_x$ & $\Delta A$ (\AA$^2$) & Proposed hypothesis \\
\midrule
\endhead
\bottomrule
\endfoot
1 & 0.01 & 400 & 0.30 & 3844.5 & Baseline run with mid-range knobs --- moderate strain rate (0.01 /ps), moderate temperature (400 K), and moderate crack length (0.3 $\times$ Lx). With no prior data, this establishes a reference point for the campaign. We expect measurable crack opening under these conditions without catastrophic failure. \\[2pt]
2 & 0.003 & 200 & 0.30 & 1489.2 & Iteration 1 produced $\Delta$ crack surface area = 3844.5 \AA{}² at mid-range values. To minimize crack opening, I'll reduce both strain rate (to 0.003 /ps --- slower loading gives the lattice more time to accommodate strain) and temperature (to 200 K --- less thermal activation of dislocation motion), while keeping crack\_length\_frac at 0.3. Lower strain rate and temperature should both reduce crack growth. \\[2pt]
3 & 0.001 & 200 & 0.30 & 1257.5 & Iterations 1→2 showed that reducing strain rate (0.01→0.003) and temperature (400→200 K) cut $\Delta$ crack area by 61\%. Now I'll push strain rate to the minimum (0.001 /ps) while keeping T=200 K and crack\_length\_frac=0.3 to isolate the strain rate effect. The slowest loading should give the lattice maximum time to blunt the crack tip via dislocation emission rather than brittle propagation. \\[2pt]
4 & 0.001 & 100 & 0.30 & 1201.3 & With strain rate already at the minimum (0.001 /ps), I'll now test the temperature effect by dropping to the lower bound (100 K) while keeping strain\_rate=0.001 and crack\_length\_frac=0.3. At 100 K, thermal activation is minimal --- Cu should behave more brittly, but the very low strain rate may still allow enough dislocation emission to blunt the crack. This isolates the temperature effect at the optimal strain rate. \\[2pt]
\textbf{5} & \textbf{0.001} & \textbf{100} & \textbf{0.10} & \textbf{0.0} & \textbf{The best so far is iter 4 (strain\_rate=0.001, T=100 K, crack\_frac=0.3) at 1201.3 \AA{}². A shorter initial crack (0.1 --- the minimum) should have a higher stress intensity factor threshold before propagation begins, meaning less crack growth under the same loading. Keeping strain\_rate=0.001 and T=100 K.} \\[2pt]
6 & 0.001 & 100 & 0.20 & 745.1 & Iter 5 achieved $\Delta$=0 at (strain\_rate=0.001, T=100 K, crack\_frac=0.1). Now I want to find the boundary --- can we still get $\Delta$=0 with a larger crack? Testing crack\_frac=0.2 while keeping the optimal strain\_rate=0.001 and T=100 K. If the crack still doesn't propagate at 0.2, the critical crack length for propagation at these conditions lies above 0.2. \\[2pt]
7 & 0.005 & 100 & 0.10 & 1041.0 & Iter 5 (crack\_frac=0.1, T=100 K, strain\_rate=0.001) gave $\Delta$=0, but iter 6 (crack\_frac=0.2) gave 745 \AA{}². Now testing whether the short crack (0.1) can resist a higher strain rate. Testing strain\_rate=0.005 with crack\_frac=0.1 and T=100 K. If $\Delta$ remains 0, the short crack is robust even under faster loading. \\[2pt]
8 & 0.001 & 400 & 0.10 & 59.6 & The $\Delta$=0 result (iter 5) required all three knobs at their extremes: lowest strain rate (0.001), lowest T (100 K), shortest crack (0.1). Iter 7 showed that even the short crack opens at strain\_rate=0.005. Now testing whether the short crack at the slowest strain rate can tolerate higher temperature --- T=400 K with strain\_rate=0.001 and crack\_frac=0.1. This probes whether temperature alone can trigger propagation in a short crack under slow loading. \\[2pt]
9 & 0.02 & 700 & 0.50 & 5036.8 & Exploring the worst-case corner to understand the full dynamic range: maximum strain rate (0.02), high temperature (700 K), and long crack (0.5). This should produce the largest $\Delta$ in the campaign and help quantify the sensitivity of each knob across the full range. \\[2pt]
10 & 0.002 & 100 & 0.10 & 383.7 & Testing whether a slightly higher strain rate (0.002) with the short crack and low T can still achieve $\Delta$=0 or near-zero. This narrows the critical strain rate threshold at the optimal crack length. Iter 5 (0.001) gave $\Delta$=0, iter 7 (0.005) gave 1041 \AA{}². \\[2pt]
11 & 0.001 & 700 & 0.50 & 2206.8 & Testing high-temperature, low-strain-rate, long-crack combination: T=700 K (max), strain\_rate=0.001 (min), crack\_frac=0.5 (max). Compared to iter 4 (T=100 K, same strain rate, crack\_frac=0.3 → 1201 \AA{}²), this reveals whether high temperature dramatically amplifies crack growth in longer cracks even at the slowest loading rate. \\[2pt]
12 & 0.02 & 700 & 0.10 & 3755.4 & Testing high strain rate with short crack at high temperature --- the opposite extreme from the winner. strain\_rate=0.02 (max), T=700 K (max), crack\_frac=0.1 (min). This tests whether crack\_length\_frac=0.1 provides any protection under the harshest conditions. \\[2pt]
13 & 0.01 & 100 & 0.50 & 5219.5 & Testing moderate strain rate (0.01), low temperature (100 K), long crack (0.5). This tests whether low temperature can protect a long crack under moderate loading --- comparing to iter 1 (same strain rate, T=400 K, crack\_frac=0.3 → 3844 \AA{}²) and iter 4 (strain\_rate=0.001, T=100 K, crack\_frac=0.3 → 1201 \AA{}²). \\[2pt]
14 & 0.005 & 400 & 0.10 & 909.2 & Testing moderate strain rate (0.005), moderate temperature (400 K), short crack (0.1). Iter 7 (0.005, 100 K, 0.1) gave 1041 \AA{}² and iter 8 (0.001, 400 K, 0.1) gave 60 \AA{}². This tests the interaction between strain rate and temperature at the short crack length. \\[2pt]
15 & 0.001 & 400 & 0.30 & 1216.0 & For the last iteration, testing low strain rate (0.001), moderate temperature (400 K), moderate crack length (0.3). This is close to iter 3 (0.001, 200 K, 0.3 → 1258 \AA{}²) but at double the temperature. It tests whether the temperature sensitivity at the optimal strain rate is linear or nonlinear when crack length is moderate. \\[2pt]
\end{longtable}

\section{Autonomous campaign: final report}
\label{sup:campaign-report}

The report below is the one ChatAtom produced at the end of the campaign, reproduced
verbatim. The objective was the change in the crack surface
area, to be minimized; the loop ran all fifteen iterations and stopped on the
iteration limit, with the best trial at iteration 5 (strain rate 0.001~ps$^{-1}$,
100~K, initial crack length 0.1 of the cell length).

This 15-iteration campaign systematically explored the crack propagation landscape in Cu under monotonic tensile loading, sweeping strain rate (0.001--0.02 /ps), temperature (100--700 K), and crack length fraction (0.1--0.5) with a fixed crack width of 4$\times$$a_0$.

\textbf{The winning configuration} --- iteration 5 (strain\_rate=0.001, T=100 K, crack\_frac=0.1) --- achieved \textbf{$\Delta$ crack surface area = 0 \AA{}$^2$}, meaning the seeded crack did not propagate at all. This was the only trial with zero crack opening across all 15 iterations.

\textbf{Strain rate was the dominant knob.} The progression from iter 1 (0.01 /ps $\rightarrow$ 3845 \AA{}$^2$) to iter 2 (0.003 $\rightarrow$ 1489 \AA{}$^2$) to iter 3 (0.001 $\rightarrow$ 1258 \AA{}$^2$) shows a strong, monotonic relationship between strain rate and crack damage, even with other knobs held constant. The critical threshold for zero propagation lies sharply between 0.001 and 0.002 /ps: iter 5 (0.001) gave $\Delta$=0 while iter 10 (0.002) gave 384 \AA{}$^2$ --- a 2$\times$ increase in strain rate broke the no-propagation regime entirely.

\textbf{Crack length fraction was the second most important knob.} At the optimal strain rate and temperature, increasing crack\_frac from 0.1 to 0.2 (iter 6) jumped $\Delta$ from 0 to 745 \AA{}$^2$, and to 0.3 (iter 4) gave 1201 \AA{}$^2$. Longer cracks have lower critical stress intensity factors, so they propagate under the same applied strain. Notably, at high strain rates, crack length mattered less --- iter 12 (0.02, 700 K, crack\_frac=0.1) still produced 3755 \AA{}$^2$ with 129 voids, showing that extreme loading overwhelms the geometric advantage of a short crack.

\textbf{Temperature had a secondary but measurable effect.} Comparing iter 3 (200 K, 0.001, 0.3 $\rightarrow$ 1258 \AA{}$^2$) vs iter 4 (100 K $\rightarrow$ 1201 \AA{}$^2$) vs iter 15 (400 K $\rightarrow$ 1216 \AA{}$^2$), the temperature sensitivity at low strain rate and moderate crack length was surprisingly weak --- all three fell in the 1200--1260 \AA{}$^2$ range. However, at the short crack length, temperature mattered more: iter 5 (100 K, 0.001, 0.1 $\rightarrow$ 0 \AA{}$^2$) vs iter 8 (400 K $\rightarrow$ 60 \AA{}$^2$) shows that thermal fluctuations can nucleate small voids even when the crack is subcritical.

\textbf{The worst-case corner} --- iter 9 (0.02, 700 K, 0.5) --- produced $\Delta$=5037 \AA{}$^2$ with 56 voids, while iter 13 (0.01, 100 K, 0.5) gave an even higher 5219 \AA{}$^2$, demonstrating that a long crack at moderate-to-high strain rate causes catastrophic damage regardless of temperature.

\textbf{Suggested next steps:} (1) A follow-up campaign with strain\_rate narrowed to [0.001, 0.003] and crack\_frac to [0.1, 0.15] would map the exact propagation threshold with higher resolution. (2) Varying crack\_width\_a0 (fixed at 4 here) could reveal whether wider cracks shift the critical strain rate. (3) A fatigue (cyclic loading) campaign at the winning conditions would test whether the no-propagation regime survives repeated loading cycles.